\documentclass[10pt,twocolumn,letterpaper]{article}

\usepackage{cvpr}              

\usepackage{makecell}
\usepackage{graphicx}

\definecolor{cvprblue}{rgb}{0.21,0.49,0.74}
\usepackage[pagebackref,breaklinks,colorlinks,allcolors=cvprblue]{hyperref}

\usepackage{booktabs}
\usepackage{arydshln}
\usepackage{amsmath}
\usepackage{amssymb}
\usepackage{adjustbox}
\usepackage{tcolorbox}
\usepackage{algorithm}
\usepackage{algpseudocode}

\def\confName{CVPR}
\def\confYear{2026}

\title{ConceptPrism: Concept Disentanglement in Personalized Diffusion Models via Residual Token Optimization}

\author{
Minseo Kim\\
KAIST, South Korea\\
{\tt\small alstj1571@kaist.ac.kr}
\and
Minchan Kwon\\
KAIST, South Korea\\
{\tt\small kmc0207@kaist.ac.kr}
\and
Dongyeun Lee\\
KAIST, South Korea\\
{\tt\small ledoye@kaist.ac.kr}
\and
Yunho Jeon\textsuperscript{\textdagger}\\
Hanbat National University, South Korea\\
{\tt\small yhjeon@hanbat.ac.kr}
\and
Junmo Kim\textsuperscript{\textdagger}\\
KAIST, South Korea\\
{\tt\small junmo.kim@kaist.ac.kr}
}

\begin{document}
\maketitle
\footnotetext{\textsuperscript{\textdagger}Joint corresponding authors.}
\begin{abstract}
    Personalized text-to-image (T2I) generation has emerged as a key application for creating user-specific concepts from a few reference images. The core challenge is concept disentanglement: separating the target concept from irrelevant residual information. Lacking such disentanglement, capturing high-fidelity features often incorporates undesired attributes that conflict with user prompts, compromising the trade-off between concept fidelity and text alignment. While existing methods rely on manual guidance, they often fail to represent intricate visual details and lack scalability. We introduce ConceptPrism, a framework that extracts shared features exclusively through cross-image comparison without external information. We jointly optimize a target token and image-wise residual tokens via reconstruction and exclusion losses. By suppressing shared information in residual tokens, the exclusion loss creates an information vacuum that forces the target token to capture the common concept. Extensive evaluations demonstrate that ConceptPrism achieves accurate concept disentanglement and significantly improves overall performance across diverse and complex visual concepts.
\end{abstract}

\vspace{-2mm}

\begin{figure}[t]
    \centering
    \resizebox{0.85\columnwidth}{!}{
        \includegraphics{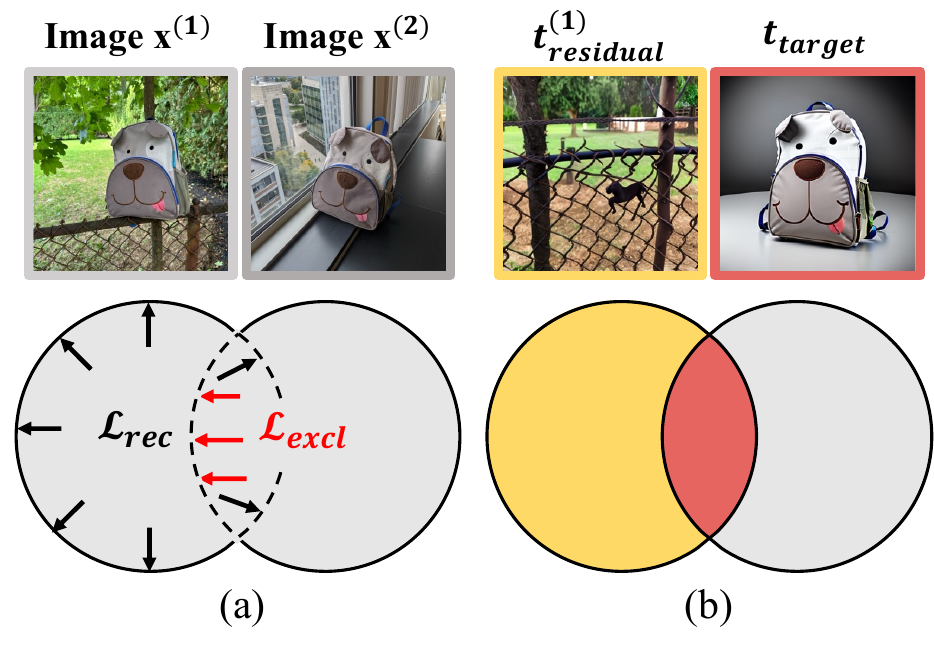}
    }
    \caption{\textbf{Motivation of ConceptPrism.} The reconstruction loss ($\mathcal{L}_{\text{rec}}$) promotes information acquisition from the given image, while the exclusion loss ($\mathcal{L}_{\text{excl}}$) compels discarding the commonalities from the set. By jointly optimizing the target and residual tokens with dual losses, we disentangle the personalized visual concept from irrelevant details without explicit guidance.}
    \vspace{-4mm}
    \label{fig:intro}
\end{figure}

\vspace{-4mm}

\section{Introduction}
\label{sec:intro}

The advent of text-to-image (T2I) diffusion models~\cite{dalle2,imagen,ldm} has opened new frontiers in digital content creation.
A key application attracting significant interest from both industry and the public is the creation of user-specific concepts like ``my doll" or ``my style"~\cite{textual_inversion,dreambooth,custom_diffusion}.
Given a handful of user-specific images, personalized generative models aim to implant a new concept into a special text token.

The core challenge of personalized T2I generation is \textbf{concept disentanglement}: separating the entangled information within the given reference images into the desired target concept and irrelevant residual information. Lacking such disentanglement, capturing more information for higher target concept fidelity inevitably incorporates undesired attributes that conflict with user text prompts, thereby degrading text alignment. This trade-off between concept fidelity and text alignment compromises the overall performance of personalized generation.

To address this, existing methods utilize manual guidance that specifies ``what to learn," such as linguistic cues~\cite{textual_inversion,dreambooth,custom_diffusion, disenbooth, disenvisioner}, segmentation masks~\cite{wei2023elite}, or auxiliary image sets~\cite{lego}. While effective, their reliance on manual guidance leads to incomplete disentanglement, as the guidance often fails to precisely represent all visual details of the target concept. Furthermore, the cumbersome framework of injecting external guidance limits the types of concepts that can be applied.

To overcome these limitations, we introduce \textbf{ConceptPrism}, a novel framework inspired by the human ability to identify common features through visual comparison (Figure~\ref{fig:intro}). Our method extracts the target concept exclusively through cross-image comparison, eliminating the need for external information. This independence from manual guidance allows for the precise extraction of highly complex visual features. Furthermore, our approach requires no auxiliary encoders to incorporate external guidance, ensuring high scalability across diverse model architectures.

Specifically, we define a target token to capture the shared concept and a set of image-wise residual tokens to explicitly provide space for irrelevant, image-specific information. These tokens are jointly optimized via two complementary objectives. First, a \textbf{reconstruction loss} ensures that the combination of the target token and a residual token can faithfully represent each corresponding reference image. Second, an \textbf{exclusion loss} is applied to the residual tokens to suppress any features shared across the image set. 
This constraint creates an information vacuum that forces the target token to capture the common concept to satisfy the reconstruction objective, providing a technical solution to identify shared concepts without manual guidance.

Our contributions are as follows:

\begin{itemize}
    \item We introduce \textbf{ConceptPrism}, a novel paradigm that utilizes cross-image comparison to disentangle concepts in personalized generation without external guidance.
    \item We propose an exclusion loss as a technical solution for identifying shared concepts, providing a mechanism that automatically isolates target features by constraining the content of residual tokens.
    \item Extensive experiments confirm that our method achieves a significantly improved generation performance, proving that ours achieves accurate concept disentanglement.
\end{itemize}

\section{Related Works}
\label{sec:relatedworks}

\paragraph{Personalized Text-to-Image (T2I) Generation.}

Personalized T2I generation aims to enable generative models to create images of a user-specific concept learned from a few references. Existing methods are broadly categorized into encoder-based and optimization-based approaches.

Encoder-based methods~\cite{xiao2024fastcomposer,hyperdreambooth,shi2024instantbooth,t2i_adapter,ipadapter,ominicontrol,syncd,ssrencoder} employ specialized encoders to extract target features from reference images, enabling personalized generation in a single shot. Many of these design domain-specific encoders for faces~\cite{xiao2024fastcomposer, hyperdreambooth}, styles~\cite{shi2024instantbooth, t2i_adapter}, or human poses~\cite{ipadapter}. Recent techniques~\cite{ominicontrol,syncd} further improve fidelity by leveraging the structural properties of DiT~\cite{dit} backbones. While computationally efficient, these approaches often lack a deep understanding of target concept attributes, leading to poor performance when prompts require significant appearance changes. Additionally, they struggle to aggregate information from multiple references into a single, robust concept.

In contrast, optimization-based approaches~\cite{textual_inversion, dreambooth, custom_diffusion} aim to create a durable model that persistently retains a user's concept, ensuring high consistency across multiple generations. Textual Inversion~\cite{textual_inversion} optimizes only a new token embedding while keeping the model frozen. DreamBooth~\cite{dreambooth} fine-tunes the entire model to capture intricate details. Subsequent studies~\cite{han2023svdiff,tewel2023key,blora,ziplora} demonstrate that effective optimization is achievable by updating only select layers or utilizing Low-Rank Adaptation (LoRA)~\cite{hu2022lora}. Following this line of work, we adopt a lightweight training scheme that fine-tunes the text tokens and the corresponding cross-attention layers via LoRA.

\vspace{-4mm}

\paragraph{Disentanglement in Personalized Generation.}

Numerous studies~\cite{tokenverse,wei2023elite,disenbooth,disenvisioner} emphasize the importance of concept disentanglement in personalized generation to accurately extract intended visual features. Existing approaches typically employ explicit guidance to facilitate this separation. ELITE~\cite{wei2023elite} utilizes a pre-defined feature hierarchy from a CLIP image encoder to isolate specific visual information. DisenBooth~\cite{disenbooth} employs learnable masks to remove regions corresponding to a class noun, while DisEnvisioner~\cite{disenvisioner} uses a dedicated tokenizer to decompose images into class-specific and residual embeddings. Other lines of work~\cite{lego,paircustomization} utilize image pairs with and without the target concept to guide the disentanglement process.

Departing from these paradigms that require pre-defined criteria, we propose a novel disentanglement paradigm that automatically identifies shared features across images as the target concept. Because our method relies solely on cross-image comparison without external guidance, it precisely captures complex visual details that manual labels or other modalities often fail to articulate.

\vspace{-1mm}
\section{Preliminaries}
\label{sec:preliminaries}

\begin{figure*}
  \centering
  \includegraphics[width=1.0\textwidth]{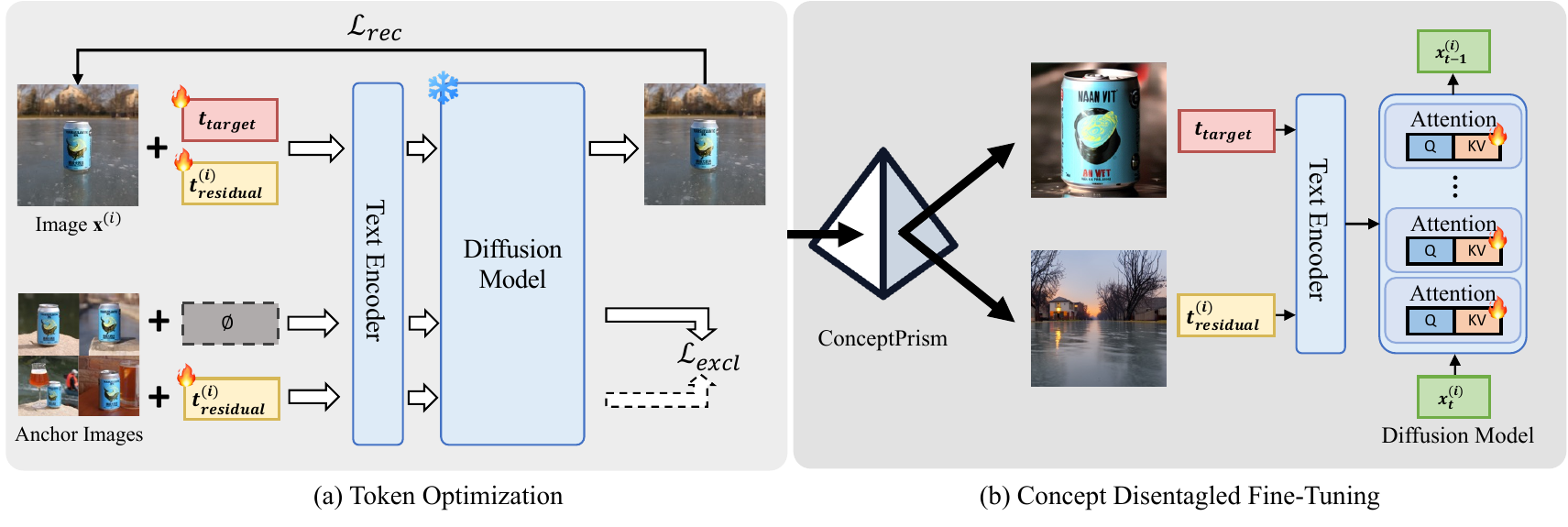}
    \vspace{-6mm}
  \caption{\textbf{Training Pipeline of ConceptPrism.} Our method comprises two stages: (a) In the token optimization, the target and image-wise residual tokens are jointly optimized via dual losses. The reconstruction loss ($\mathcal{L}_{\text{rec}}$) guides the faithful reconstruction of the given image by conditioning on both tokens simultaneously. The exclusion loss ($\mathcal{L}_{\text{excl}}$) forces the residual token to be uninformative of the shared target concept $\mathcal{C}_{\text{target}}$ by matching the unconditional generation probability distribution. (b) In the subsequent fine-tuning stage, the learned tokens initialize the model to focus only on the necessary concept, effectively resolving the trade-off caused by concept entanglement.}
  \label{fig:method}
    \vspace{-2mm}
\end{figure*}

\subsection{Text-to-Image (T2I) Diffusion Models}

Given a large-scale image dataset, diffusion models $p_{\theta}$ approximate the true data distribution $p_{data}$ via a tractable noise matching process~\cite{ddpm, song2020score}. Specifically, Stable Diffusion~\cite{ldm}, the model used in our study, employs a VAE~\cite{vae} encoder $\mathcal{E}(\cdot)$ to transform an input image $\mathbf{x}$ into a low-dimensional latent representation $\mathbf{z}=\mathcal{E}(\mathbf{x})$.
Under the DDPM framework~\cite{ddpm}, the task of approximating $p_{\theta}$ simplifies to a noise prediction problem. Given a noisy latent $\mathbf{z}_{t}$ obtained by progressively adding noise $\epsilon \sim \mathcal{N}(0, \mathbf{I})$ to $\mathbf{z}$ according to a timestep $t \sim \mathcal{U}([0,1])$, the noise prediction model $\epsilon_{\theta}$ is trained with the following objective:

\vspace{-3mm}

\begin{equation}
    \mathcal{L}_{\text{DM}} = \mathbb{E}_{\mathbf{z}, c, \epsilon,t}\left[w_{t}\left\| \epsilon - \epsilon_{\theta}(\mathbf{z}_{t}, t, \Gamma(c) \right)\|_{2}^{2}\right],
    \label{eq:dm_loss}
\end{equation}

\noindent where $\Gamma(\cdot)$ denotes a text encoder such as CLIP~\cite{clip}, and $w_{t}$ is a weighting term defined by the noise schedule. The text condition $c$ is injected into the cross-attention layers of the denoising model $\epsilon_{\theta}$ to control the image generation process.

\subsection{Personalized T2I Generative Models}

Given a small set of user-specific images $\{\mathbf{x}^{(i)} \}_{i=1}^{N}$, we formally decompose the visual information in each image $\mathbf{x}^{(i)}$ into two parts: $\{\mathcal{C}_{\text{target}}, \mathcal{C}_{\text{residual}}^{(i)}\}$. The \textbf{target concept} $\mathcal{C}_{\text{target}}$ represents the shared visual features across all $N$ images, while the image-wise \textbf{residual concept} $\mathcal{C}_{\text{residual}}^{(i)}$ comprises the remaining image-specific elements unique to each $\mathbf{x}^{(i)}$.

Existing approaches typically aim to map $\mathcal{C}_{\text{target}}$ into a single target text token $\mathbf{t}_{\text{target}}$. This is achieved by optimizing either the token~\cite{textual_inversion} or the model parameters~\cite{dreambooth,custom_diffusion} to reconstruct the $N$ images using the diffusion model loss, conditioned on $\mathbf{t}_{\text{target}}$. However, the reconstruction objective inevitably forces the token to also capture the residual information, causing $\mathcal{C}_{\text{residual}}^{(i)}$ to become entangled with $\mathbf{t}_{\text{target}}$.

\vspace{-2mm}

\section{Our Approach: ConceptPrism}
\label{sec:methods}

\subsection{Image-wise Residual Tokens}
\label{sec:3.1}

To address the challenge of concept disentanglement, we introduce \textbf{ConceptPrism}, a framework that captures the target visual concept $\mathcal{C}_{\text{target}}$ through cross-image comparison. To represent both shared and image-specific features, we assign distinct text tokens to capture each disentangled concept: a single \textbf{target token} $\mathbf{t}_{\text{target}}$ for $\mathcal{C}_{\text{target}}$ and a set of \textbf{residual tokens} $\{\mathbf{t}_{\text{residual}}^{(i)}\}_{i=1}^{N}$ for the image-specific elements $\{\mathcal{C}_{\text{residual}}^{(i)}\}_{i=1}^{N}$.

When training on the $i$-th image $\mathbf{x}^{(i)}$, we provide a text condition $c^{(i)}$ formatted as ``$\mathbf{t}_{\text{target}}$ with $\mathbf{t}_{\text{residual}}^{(i)}$". Our reconstruction loss $\mathcal{L}_{\text{rec}}$ then optimizes these tokens to ensure their combination faithfully represents $\mathbf{x}^{(i)}$, formulated as:

\vspace{-4mm}
\begin{equation}
    \mathcal{L}_{\text{rec}} = \sum_{i=1}^N\ \underset{\epsilon,t}{\mathbb{E}}\left[w_{t}\left\| \epsilon - \epsilon_{\theta}(\mathbf{z}^{(i)}_{t}, t, \Gamma(c^{(i)})) \right\|_{2}^{2}\right].
    \label{eq:rec_loss}
\end{equation}
\vspace{-5mm}

\begin{table*}[htbp]
\centering
\caption{\textbf{Quantitative Evaluations on the DreamBench.} CLIP-T measures text alignment; DINO / CLIP-I assess concept fidelity. Training time to the best quality step was also measured (N/A for tuning-free approaches ELITE and DisEnvisioner). All experiments were repeated over 4 different seeds. Ours achieves the highest performance across all three metrics with negligible computational overhead.}
\vspace{-2mm}
\begin{tabular*}{\textwidth}{@{\extracolsep{\fill}} lcccc}
    \toprule
    Method & CLIP-T $\uparrow$ & DINO $\uparrow$ & CLIP-I $\uparrow$ & Training Time(s) $\downarrow$ \\
    \midrule
    Pretrained SD~\cite{ldm} & 0.377 \scriptsize$\pm$0.024 & 0.167 \scriptsize$\pm$0.029 & 0.320 \scriptsize$\pm$0.030 & - \\
    DreamBooth-LoRA~\cite{dreambooth} & 0.313 \scriptsize$\pm$0.030 & \underline{0.182} \scriptsize$\pm$0.030 & 0.331 \scriptsize$\pm$0.030 & 434.32 \scriptsize$\pm$75.50 \\
    Custom Diffusion~\cite{custom_diffusion} & \textbf{0.357} \scriptsize$\pm$0.023 & 0.178 \scriptsize$\pm$0.032 & \underline{0.341} \scriptsize$\pm$0.031 & 459.60 \scriptsize$\pm$59.64 \\
    DisenBooth~\cite{disenbooth} & \underline{0.355} \scriptsize$\pm$0.027 & 0.170 \scriptsize$\pm$0.035 & 0.324 \scriptsize$\pm$0.033 & 928.63 \scriptsize$\pm$51.43 \\
    DisEnvisioner~\cite{disenvisioner} & 0.309 \scriptsize$\pm$0.023 & 0.171 \scriptsize$\pm$0.022 & 0.320 \scriptsize$\pm$0.024 & - \\
    ELITE~\cite{wei2023elite} & 0.293 \scriptsize$\pm$0.030 & 0.172 \scriptsize$\pm$0.031 & \underline{0.341} \scriptsize$\pm$0.032 & - \\
    \midrule
    ConceptPrism (Ours) & \textbf{0.357} \scriptsize$\pm$0.025 & \textbf{0.210} \scriptsize$\pm$0.035 & \textbf{0.353} \scriptsize$\pm$0.032 & 477.30 \scriptsize$\pm$91.29 \\
    \bottomrule
\end{tabular*}
\label{tab:main}
\vspace{-5mm}
\end{table*}

\subsection{Exclusion Loss for Concept Disentanglement}
\label{sec:3.2}

Alongside the reconstruction loss, we introduce an exclusion loss $\mathcal{L}_{\text{excl}}$ to robustly disentangle concepts. Relying solely on $\mathcal{L}_{\text{rec}}$ causes the residual token $\mathbf{t}_{\text{residual}}^{(i)}$ to overfit by memorizing its corresponding image $\mathbf{x}^{(i)}$. Such overfitting leads to a trivial solution where the target token $\mathbf{t}_{\text{target}}$ converges to an uninformative state, failing to capture the shared features across the image set.

To prevent the target concept from being absorbed into the residual tokens, $\mathcal{L}_{\text{excl}}$ guides the residual tokens to avoid capturing $\mathcal{C}_{\text{target}}$. From the perspective of diffusion-based generative models, this implies that the condition $\mathbf{t}_{\text{residual}}^{(i)}$ should be uninformative for reconstructing the distribution of $\mathcal{C}_{\text{target}}$. We thus design the exclusion loss as the KL divergence between two distributions: the reconstruction distribution of $\mathcal{C}_{\text{target}}$ conditioned on $\mathbf{t}_{\text{residual}}^{(i)}$, and that under the null condition $\varnothing$. By minimizing this divergence, the residual token is compelled to discard $\mathcal{C}_{\text{target}}$.

As $\mathcal{C}_{\text{target}}$ is defined abstractly, we utilize images $\mathbf{x}^{(j)}$ ($j \neq i$) as a proxy for the target concept distribution for each $\mathbf{t}_{\text{residual}}^{(i)}$. Since the diffusion model defines a generative process as a Markov chain of iterative reverse steps, our objective is formulated as:

\vspace{-4mm}

\begin{equation}
    \begin{split}
        \underset{\{\mathbf{t}_{\text{residual}}^{(k)}\}_{k=1}^N}{\text{min}}\ \mathbb{E}_{i, \mathbf{z}_t^{(j\neq i)}, t} \left[ D_{\text{KL}} \left( p_{\theta}(\mathbf{z}_{t-1}^{(j)} | \mathbf{z}_{t}^{(j)}, t, \varnothing) \right. \right. \\ \left. \left. \parallel p_{\theta}(\mathbf{z}_{t-1}^{(j)} | \mathbf{z}_{t}^{(j)}, t, \Gamma(c_{\text{residual}}^{(i)})) \right) \right],
    \end{split}
\end{equation}

\vspace{-1mm}
where $c_{\text{residual}}^{(i)}$ is the text condition containing only the $i$-th residual token (e.g., ``$\mathbf{t}_{\text{residual}}^{(i)}$"). This objective can be simplified into the following tractable noise matching process, as derived in Appendix.

\vspace{-4mm}

\begin{equation}
\label{eq:excl_loss}
\mathcal{L}_{\text{excl}} \!=\! \mathbb{E}_{i,j\neq i, \epsilon, t} \! \left[w_{t} \! \left\| \epsilon_\theta(\mathbf{z}_{t}^{(j)}, t,\Gamma(c_{\text{residual}}^{(i)})) \!-\! \epsilon_\theta(\mathbf{z}_{t}^{(j)}, t,\varnothing) \right\|_2^2\right].
\end{equation}

Notably, the index of the image latent $\mathbf{z}_{t}^{(j)}$ differs from that of the residual text condition $c_{\text{residual}}^{(i)}$,  which constitutes the key idea behind this loss. Our final objective consists of dual losses, where hyperparameter $\beta$ controls their relative importance:

\vspace{-1mm}

\begin{equation}
    \mathcal{L}_{\text{total}} = \mathcal{L}_{\text{rec}} + \beta\ \mathcal{L}_{\text{excl}}.
    \label{eq:total_loss}
\end{equation}

\vspace{2mm}

\begin{figure}
  \centering
  \includegraphics[width=0.5\textwidth]{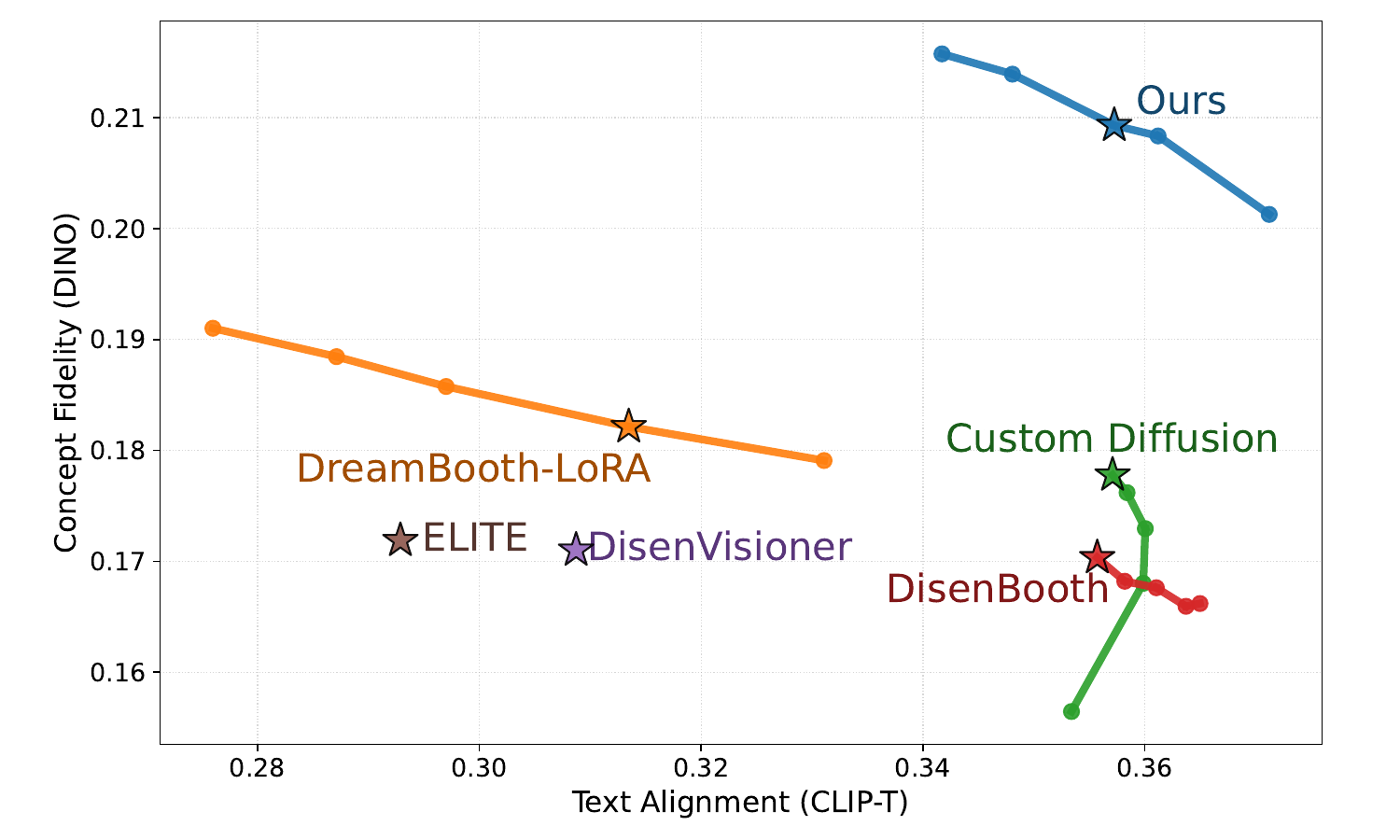}
  \caption{\textbf{Performance trade-off by training steps.} Our performance was measured at 40-step intervals, others at 200-step intervals. Asterisk($\star$) denotes the best visual quality step for each method. Ours demonstrates a superior trade-off over all baselines, highlighting its effective concept disentanglement.}
  \vspace{-6mm}
  \label{fig:main_tradeoff}
\end{figure}

\vspace{-4mm}

\subsection{Better Initialization of Residual Tokens}
\label{sec:3.3}

Without careful residual token initialization, the concept disentanglement can be incomplete, with parts of the residual concept remaining entangled in $\mathbf{t}_{\text{target}}$. To ensure the residual token $\mathbf{t}_{\text{residual}}^{(i)}$ comprehensively captures information other than $\mathcal{C}_{\text{target}}$, we initialize it to contain general information about the image $\mathbf{x}^{(i)}$. We feed a descriptive sentence (8-32 words) describing the whole scene of the given image into the CLIP encoder and use its mean embeddings as the starting point for the residual token.

Interestingly, the target token $\mathbf{t}_{\text{target}}$ does not require specific prior information for initialization. Despite starting from a random point, $\mathbf{t}_{\text{target}}$ successfully captures the intended concept. During training, inter-image comparison compels the residual tokens to discard the shared concept. This process creates an information vacuum within the residual tokens, which the target token subsequently fills to accurately absorb $\mathcal{C}_{\text{target}}$ without direct supervision.

\vspace{-2mm}

\subsection{Concept Disentangled Fine-Tuning}
\label{sec:3.4}
The token optimization stage precisely isolates the intended concept into the target token. By encapsulating ``what to learn" exactly, the target token enables diffusion models to focus on the necessary information. In the fine-tuning stage, we train the model on reference images using $\mathcal{L}_{rec}$, providing the learned tokens as text conditions. The image-wise residual token is also provided alongside the target token, preventing unnecessary information from becoming associated with the target token. We insert LoRA~\cite{hu2022lora} to modify only the attention layers of the diffusion model for parameter-efficient fine-tuning, as previous literatures~\cite{disenbooth,park2025steering} have demonstrated that modifying only the attention layers retains sufficient expressiveness.

\vspace{-2mm}

\section{Experiments}
\label{sec:experiments}

\begin{figure*}
  \centering
  \includegraphics[width=0.95\textwidth]{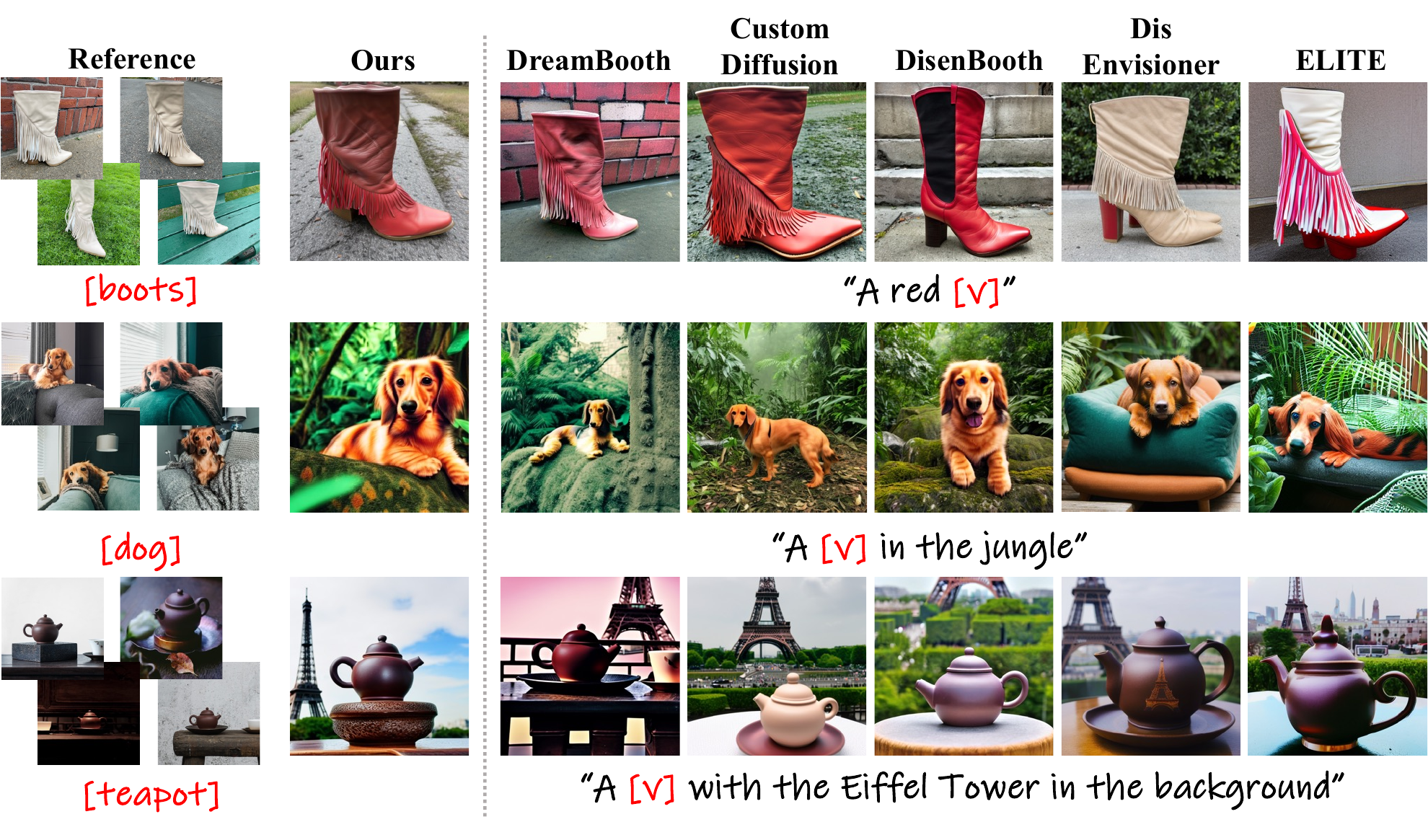}
  \vspace{-4mm}
  \caption{\textbf{Qualitative Results on various subjects.} Our method showed balanced performance, maintaining the visual details of the personalized subject while adhering to the text prompt. Other baselines tend to either copy input images (DreamBooth) or ignore concept representation to follow the prompt (Custom Diffusion, DisenBooth).}
  \label{fig:qual_object}
  \vspace{-4mm}
\end{figure*}

\subsection{Experimental Setup}

\paragraph{Datasets and Evaluations.}

We benchmark our method on DreamBench~\cite{dreambooth}, comprising 30 subjects with 4--6 images and 25 prompts each. We generate four images per prompt, totaling 3,000 images per method. Qualitative analysis further includes CustomConcept101~\cite{custom_diffusion}, DreamBench++~\cite{peng2024dreambench++}, and other open-source datasets~\cite{actionbench, Freepik}.

Performance is quantified across two dimensions: text alignment and concept fidelity. Text alignment is measured via CLIP-T~\cite{clip}, which calculates the cosine similarity between prompts and generated images. Concept fidelity is assessed using CLIP-I and DINOv2 (DINO)~\cite{oquab2023dinov2} similarity scores between reference and generated images. To isolate the target subject for accurate scoring, we apply binary masks to the reference images.

\paragraph{Implementation Details.}
For training, we set a batch size of 1 with four gradient accumulation steps in 512x512 resolution.
The learning rate was 4e-4 with the AdamW optimizer.
We set the target and residual token lengths to 1 and 8, respectively, and the exclusion loss weight $\beta$ to 0.05 by default.
For inference, we adopted the DDIM~\cite{ddim} sampler with 25 steps and a classifier-free guidance~\cite{cfg} scale of 7.5.
The token optimization stage ran for 200 steps, taking about 4 minutes on a single NVIDIA RTX 3090 GPU. The subsequent fine-tuning stage followed the same hyperparameters, achieving optimal quality after 120 steps. The performance was measured every 40 steps up to 200 steps of fine-tuning.
Built on Stable Diffusion v2.1~\cite{ldm}, our pipeline generalizes to diverse architectures including DiT~\cite{dit}.

\vspace{-5mm}

\paragraph{Baselines.}
We evaluate our method against five baselines. DreamBooth~\cite{dreambooth} and Custom Diffusion~\cite{custom_diffusion} serve as representative optimization-based methods. For a fair comparison, DreamBooth is implemented using LoRA to update only the attention layers, matching our configuration (Sec.~\ref{sec:3.4}). ELITE~\cite{wei2023elite}, DisenBooth~\cite{disenbooth}, and DisEnvisioner~\cite{disenvisioner} are recent disentanglement-focused approaches, with ELITE and DisEnvisioner being tuning-free, encoder-based models. Notably, all baselines require a class noun for the target concept, whereas our method does not. Performance is measured every 200 steps up to 1000. DreamBooth, Custom Diffusion, and DisenBooth reach optimal quality at 400, 1000, and 1000 steps, respectively.



\begin{figure*}[t]
  \centering
  \includegraphics[width=1.0\textwidth]{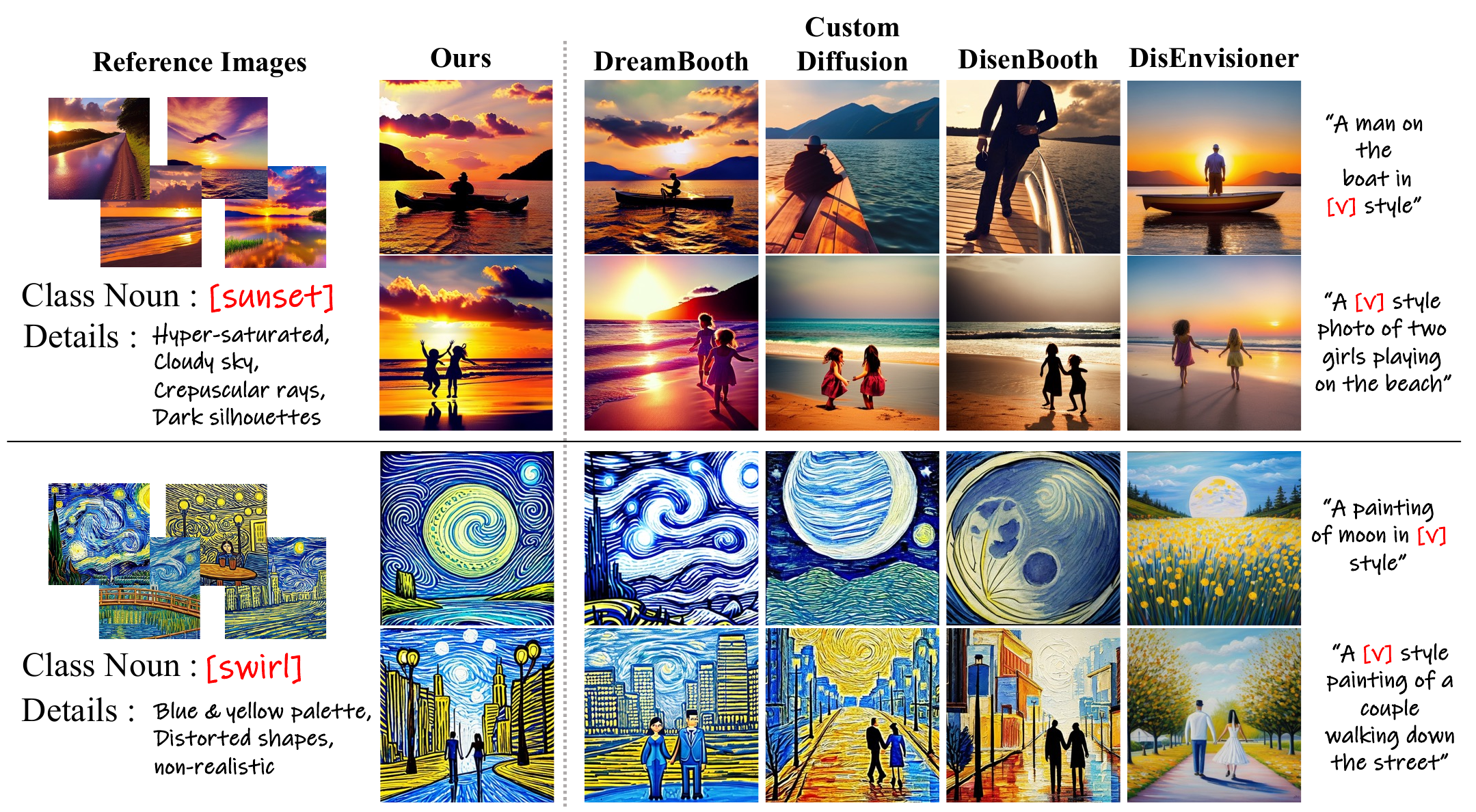}
  \caption{\textbf{Qualitative Results on abstract styles.} Our method learns the abstract style beyond a single word class noun. DreamBooth generates similar scenes due to overfitting to the reference images. Other baselines fail to capture the given visual details, adhering only to the linguistic cues (e.g., ``sunset," ``swirl").}
  \vspace{-4mm}
  \label{fig:qual_style}
\end{figure*}

\subsection{Quantitative Evaluations}
\label{sec:expr_quant}

\paragraph{Comparison with Baselines.}

Table~\ref{tab:main} reports the performance at the qualitatively optimal training steps for each method. The pre-trained SD~\cite{ldm} (before fine-tuning) serves as a baseline, representing an upper bound for CLIP-T and a lower bound for DINO and CLIP-I. Our method achieves the highest performance in both text alignment (CLIP-T) and concept fidelity (DINO, CLIP-I). While Custom Diffusion and DisenBooth show competitive CLIP-T, their DINO scores remain low. DreamBooth records a relatively high DINO score, but this comes at a significant cost to CLIP-T, suggesting overfitting to the reference images. In contrast, our method substantially improves the DINO score while maintaining high CLIP-T, effectively resolving the fidelity-alignment trade-off.

Figure~\ref{fig:main_tradeoff} further illustrates our superior trade-off. While DreamBooth, DisenBooth, and our method all exchange text alignment for fidelity during training, the differing slopes highlight that our approach sacrifices the least alignment for its fidelity gain. Furthermore, our high initial DINO scores (above 0.2) prove the effectiveness of our token initialization. Custom Diffusion maintains stable text alignment, but its fidelity plateaus after 1000 steps. Encoder-based methods, DisEnvisioner and ELITE, perform poorly on both metrics, confirming the limited precision of encoder-based techniques in concept representation.

\paragraph{Training Time.}
We compare training efficiency across methods. Unlike existing approaches that rely on a single fine-tuning stage, our method adds a 200-step token optimization process. However, this optimization significantly reduces the number of model fine-tuning steps required to reach high concept fidelity. For instance, our method achieves optimal quality in only 120 steps, compared to 400 for DreamBooth. As shown in Table~\ref{tab:main}, we are approximately twice as fast as DisenBooth and maintain minimal overhead relative to DreamBooth and Custom Diffusion given the substantial performance gains.


\subsection{Qualitative Analysis}

\paragraph{Discrete Subject Personalization.}

As visualized in Figure~\ref{fig:qual_object}, our method maintains high subject fidelity while strictly adhering to prompts. In contrast, DreamBooth often entangles background elements (e.g., red bricks) and replicates reference compositions. Our selective transformation of the subject demonstrates that token separation effectively disentangles the target concept. While Custom Diffusion and DisenBooth offer high text alignment, they lack fidelity and often ignore the subject in complex prompts. Conversely, our method preserves intricate details even under challenging conditions. Tuning-free methods, such as DisEnvisioner and ELITE, show limited conceptual understanding.

\begin{table}[h]
  \centering
  \caption{\textbf{Ablation Studies on Token Optimization.} We measured performance by ablating each component of the token optimization stage. The introduction of residual tokens and the exclusion loss significantly improved CLIP-T and DINO, respectively, verifying their role in concept disentanglement.}
  \vspace{-2mm}
   \resizebox{1.0\columnwidth}{!}{
   \begin{tabular}{lccc}
    \toprule
                            & CLIP-T $\uparrow$ & DINO $\uparrow$ \\
    \midrule
    Target Token Optimization               & 0.345                & 0.197              \\
    + Residual Tokens (Sec~\ref{sec:3.1})           & \textbf{0.358}       & 0.183              \\
    + Exclusion Loss (Sec~\ref{sec:3.2})            & 0.354                & 0.207              \\
    \midrule
    + \makecell{Better Initialization (Sec~\ref{sec:3.3})} & 0.357 & \textbf{0.210} \\
    \bottomrule
  \end{tabular}
  }
  
  \label{tab:ablation}
    \vspace{-0mm}
\end{table}

\paragraph{Abstract Style Personalization.}

Our approach also excels at learning abstract styles (Figure~\ref{fig:qual_style}). While Custom Diffusion and DisEnvisioner follow class nouns, they fail to replicate the visual essence of reference images, highlighting the limitations of linguistic cues for complex styles. By comparing features across the image set, our method captures subtle visual details that are difficult to describe with text. Notably, only ConceptPrism comprehensively reflects the intricate details of the provided styles. DreamBooth captures some stylistic elements but often struggles with creativity, frequently reproducing reference compositions. ELITE is excluded here as it requires binary masks for the target subject.

\begin{table}[h]
  \centering
  \caption{\textbf{Performance by Residual Token Length.} The presence of residual tokens ($n \geq 1$) creates a significant performance gap in both criteria, demonstrating their key role in resolving concept entanglement. The stable performance from $n=1$ to $8$ indicates that our framework is robust to token length.}
  \vspace{-2mm}
    \begin{tabular}{cccccc} 
      \toprule
      $n$ & 0 & 1 & 2 & 4 & 8 \\
      \midrule
      CLIP-T & 0.345 & 0.354 & 0.357 & 0.355 & 0.357 \\
      DINO   & 0.197 & 0.209 & 0.204 & 0.207 & 0.210 \\
      \bottomrule
    \end{tabular}
  \vspace{-2mm}
  \label{tab:res_token_length}
\end{table}
\begin{figure}
  \centering
  \includegraphics[width=1.0\columnwidth]{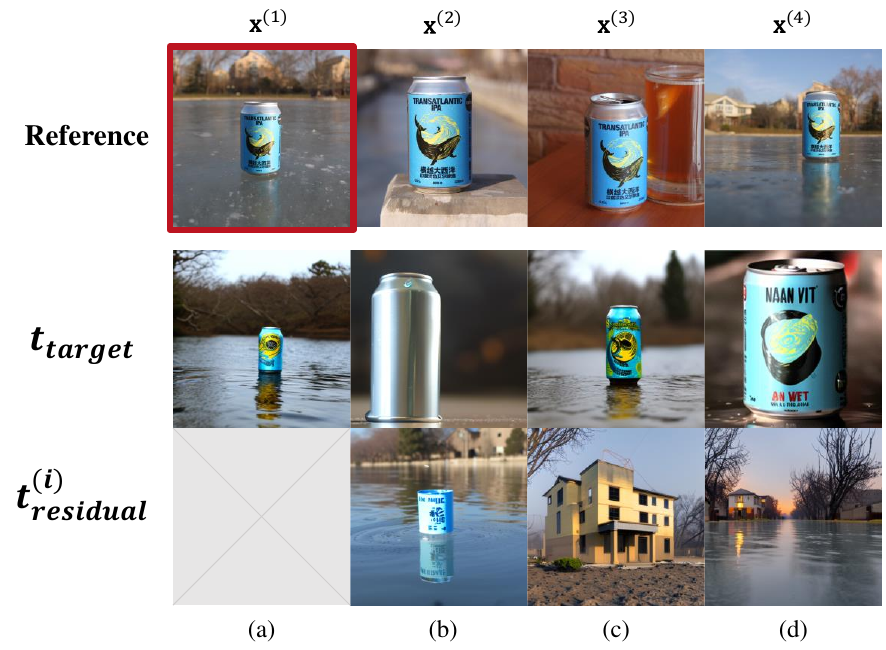}
  \vspace{-6mm}
  \caption{\textbf{Visualizations of Tokens after Token Optimization.} \textbf{(a)} Optimizing the target token alone. \textbf{(b)} Jointly optimizing target and residual tokens using the single objective $\mathcal{L}_{rec}$. \textbf{(c)} Optimizing with dual objectives by introducing $\mathcal{L}_{excl}$. \textbf{(d)} Initializing residual tokens with image descriptions.}
  \vspace{-4mm}
  \label{fig:ablation_main}
\end{figure}

\section{Ablation Studies}
\label{sec:ablation}

\paragraph{Ablations on Token Optimization.}

We investigate the impact of each component introduced in the token optimization process on concept disentanglement. Table~\ref{tab:ablation} and Figure~\ref{fig:ablation_main} shows the results of the ablation experiment conducted under the same settings as the main experiments. \textbf{(a)} Optimizing the target token alone inevitably leads to concept entanglement. \textbf{(b)} When a residual token is added and optimized with $\mathcal{L}_{rec}$, the residual concept is effectively separated, leading to a significant improvement in text alignment. However, the target token fails to fully capture the common features across the given image set. \textbf{(c)} Adding $\mathcal{L}_{excl}$ effectively guides the residual token to discard these common features. Consequently, the target token fully captures the target concept, leading to concept fidelity improvement. \textbf{(d)} Finally, initializing the residual token with image descriptions separates the concepts into the intended commonalities and differences, enhances both criteria.

\vspace{-3mm}


\begin{table}[h]
  \centering
  \caption{\textbf{Influence of Residual Token Initialization.} Performance remains robust across varying description lengths, demonstrating that our method is insensitive to the initial point of the residual tokens. The significant performance drop in the frozen setting confirms that our improvements stem from joint token optimization rather than the quality of the initial descriptions.}
  \vspace{-1mm}
    \begin{tabular*}{\columnwidth}{@{\extracolsep{\fill}}lcc}
        \toprule
        Description Length & CLIP-T $\uparrow$ & DINO $\uparrow$ \\
        \midrule
        Long (frozen) & 0.313 & 0.197 \\
        \midrule
        Short (1-4 words) & 0.353 & 0.198 \\
        Medium (4-8 words) & 0.356 & 0.200 \\
        Long (Ours, 8-32 words) &\textbf{ 0.357} & \textbf{0.210} \\
        \bottomrule
    \end{tabular*}
  \vspace{-2mm}
  \label{tab:res_token_init}
\end{table}

\paragraph{Influence of Residual Token Length.}

Table~\ref{tab:res_token_length} reports quantitative evaluations as the residual token length $n$ is varied. Both CLIP-T and DINO scores improve significantly when $n \geq 1$, confirming that assigning explicit space for residual information is crucial for concept disentanglement.

Results for $n \in [1, 8]$ are consistent, indicating that even a single token provides sufficient capacity for residual features. Moreover, the stability observed as $n$ increases confirms that the exclusion loss effectively prevents residual tokens from absorbing the target concept. Consequently, our method is highly robust to $n$, requiring minimal hyperparameter tuning for diverse concepts.


\begin{figure}
  \centering
  \includegraphics[width=1.0\columnwidth]{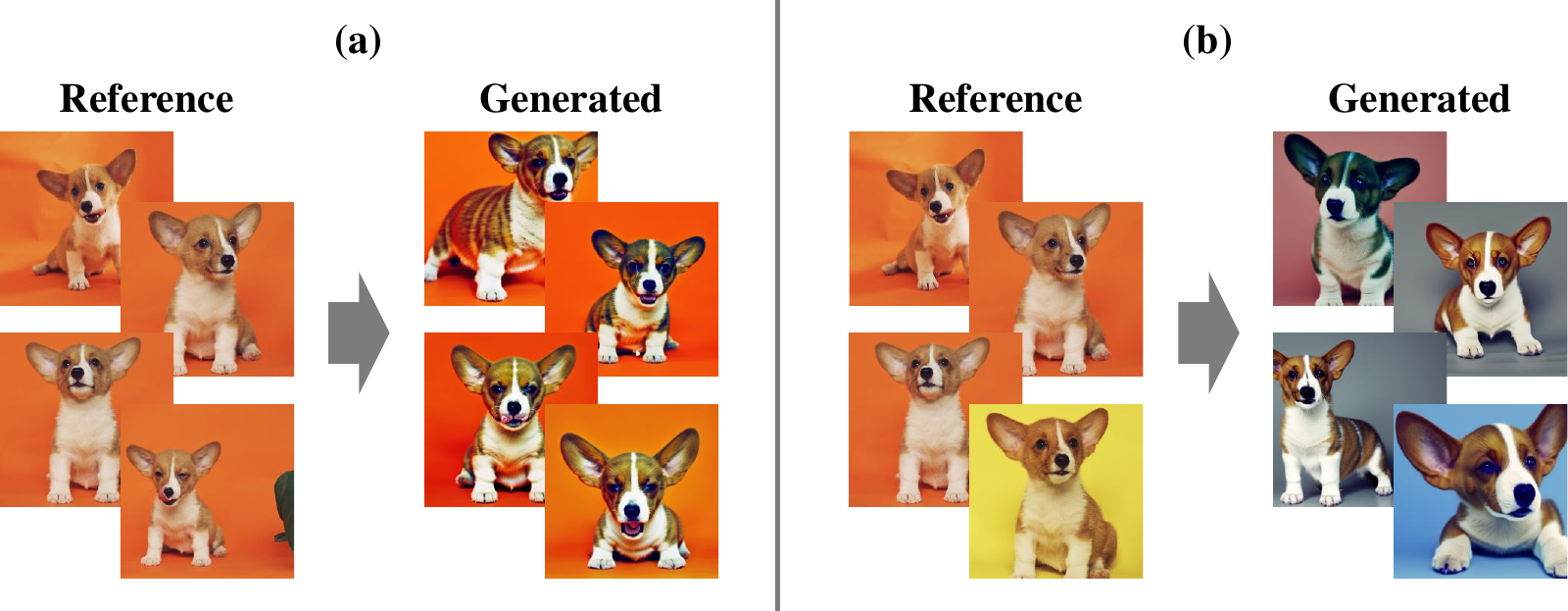}
  \vspace{-4mm}
  \caption{\textbf{Robustness on the reference set diversity.} \textbf{(a)} When all reference images share unintended visual features, such as an orange background, these elements can become entangled with the target concept. \textbf{(b)} Adding a single counterexample with a different background color effectively prevents such unintended features from being captured, ensuring accurate concept disentanglement.}
  \vspace{-2mm}
  \label{fig:orange_dog}
\end{figure}

\paragraph{Influence of Residual Token Initialization.}

Table~\ref{tab:res_token_init} analyzes the impact of residual token initialization. We utilize VLM-generated image descriptions (8--32 words) as initial points for the residual tokens, with further details provided in the Supplementary. Shortening descriptions to medium (4--8 words) or short (1--4 words) length results in minor CLIP-T and DINO drops due to lost visual details, but the overall impact remains non-decisive. Such stability suggests our method is insensitive to the specific initial point of residual tokens. Furthermore, as shown in the first row of Table~\ref{tab:res_token_init}, relying on fixed (frozen) residual tokens causes CLIP-T to drop significantly to DreamBooth levels. This highlights that our superior performance stems from the joint training of tokens rather than the descriptive quality of the initial text.

\vspace{-4mm}
\begin{figure*}[t]
  \centering
  \includegraphics[width=1.0\textwidth]{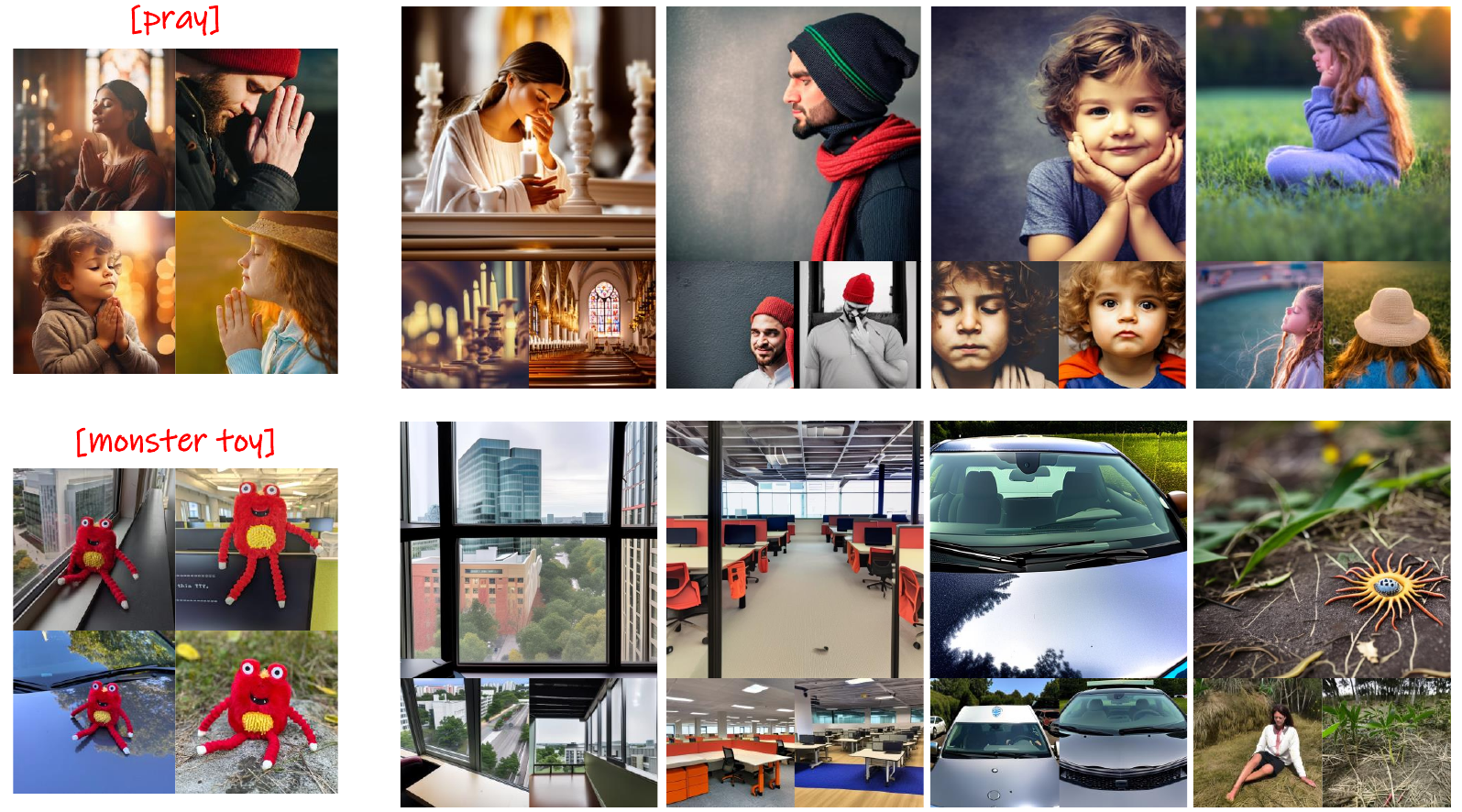}
  \vspace{-6mm}
  \caption{\textbf{Visualization of Learned Residual Tokens.} After the token optimization stage, an image-wise residual token can be solely conditioned to generate scene details without the target concept. We visualized the residual tokens learned from each reference image using three different seeds. The ability to generate diverse images of the same semantics (e.g., varied office layouts) rather than mere copies demonstrates that each token has learned an abstract residual concept.}
  \label{fig:res_token_visualization}
  \vspace{-5mm}
\end{figure*}

\paragraph{Robustness on the reference set diversity.}

Existing methods often fail to disentangle concepts when manual guidance is underspecified. Similarly, our method may capture unintended commonalities if the reference set shares visual features beyond the target concept. As shown in Figure~\ref{fig:orange_dog}a, if all images share a common visual feature (orange background) other than the target concept (dog), this unintended information can become entangled. This issue is easily mitigated by introducing a counterexample. As illustrated in Figure~\ref{fig:orange_dog}b, adding even a single image with a different background enables the exclusion loss to isolate the background as a residual feature, ensuring accurate concept disentanglement.

\vspace{-4mm}

\paragraph{Visualizations of Residual Tokens.}

To verify if our residual tokens truly learn image-wise residual information, we visualized the residual tokens after token optimization as shown in Figure~\ref{fig:res_token_visualization}. The learned residual token was used for T2I generation with the template ``A photo of $\mathbf{t}_{\mathrm{residual}}$". When the target is a discrete object, the residual tokens represent backgrounds, lighting, and camera angles while excluding the object itself. For poses, the residual token generates landscapes or subjects with the target pose removed. The ability to generate diverse images with the same semantics (e.g., offices with various layouts and arrangements) rather than simply copying the given images proves that the residual tokens capture high-level image semantics instead of merely retaining their initial values.

\vspace{-3mm}

\section{Conclusions}
\label{sec:conclusion}

In this paper, we introduced ConceptPrism, a novel concept disentanglement paradigm in personalized text-to-image generation. Unlike existing approaches that rely on manual guidance, our method automatically disentangles the shared visual concept from image-specific residuals through cross-image comparison. By jointly optimizing target and residual tokens with complementary reconstruction and exclusion objectives, the target token captures the detailed visual concept without explicit supervision. Extensive experiments demonstrate that ConceptPrism significantly improves the trade-off between concept fidelity and text alignment, establishing a robust and broadly applicable solution for diverse personalization tasks.

\vspace{-3mm}
\section{Acknowledgements}

This work was supported by Center for Applied Research in Artificial Intelligence(CARAI) grant funded by Defense Acquisition Program Administration(DAPA) and Agency for Defense Development(ADD) (UD230017TD), and by Basic Science Research Program through the National Research Foundation of Korea(NRF) funded by the Ministry of Education (RS-2025-25432454).
{
    \small
    \bibliographystyle{ieeenat_fullname}
    \bibliography{main}

@String(TOG= {ACM Trans. Graph.})

@String(ICLR = {Int. Conf. Learn. Represent.})

@String(AAAI = {AAAI})

@String(TOG   = {ACM TOG})

@String(ICLR  = {ICLR})

@inproceedings{actionbench,
  title={Learning disentangled identifiers for action-customized text-to-image generation},
  author={Huang, Siteng and Gong, Biao and Feng, Yutong and Chen, Xi and Fu, Yuqian and Liu, Yu and Wang, Donglin},
  booktitle={Proceedings of the IEEE/CVF Conference on Computer Vision and Pattern Recognition},
  pages={7797--7806},
  year={2024}
}

@inproceedings{park2025steering,
  title={Steering Guidance for Personalized Text-to-Image Diffusion Models},
  author={Park, Sunghyun and Choi, Seokeon and Park, Hyoungwoo and Yun, Sungrack},
  booktitle={Proceedings of the IEEE/CVF International Conference on Computer Vision},
  pages={15907--15916},
  year={2025}
}

@inproceedings{lego,
  title={Lego: Learning to disentangle and invert personalized concepts beyond object appearance in text-to-image diffusion models},
  author={Motamed, Saman and Paudel, Danda Pani and Van Gool, Luc},
  booktitle={European Conference on Computer Vision},
  pages={116--133},
  year={2024},
  organization={Springer}
}

@inproceedings{tewel2023key,
  title={Key-locked rank one editing for text-to-image personalization},
  author={Tewel, Yoad and Gal, Rinon and Chechik, Gal and Atzmon, Yuval},
  booktitle={ACM SIGGRAPH 2023 conference proceedings},
  pages={1--11},
  year={2023}
}

@inproceedings{shi2024instantbooth,
  title={Instantbooth: Personalized text-to-image generation without test-time finetuning},
  author={Shi, Jing and Xiong, Wei and Lin, Zhe and Jung, Hyun Joon},
  booktitle={Proceedings of the IEEE/CVF conference on computer vision and pattern recognition},
  pages={8543--8552},
  year={2024}
}

@inproceedings{t2i_adapter,
  title={T2i-adapter: Learning adapters to dig out more controllable ability for text-to-image diffusion models},
  author={Mou, Chong and Wang, Xintao and Xie, Liangbin and Wu, Yanze and Zhang, Jian and Qi, Zhongang and Shan, Ying},
  booktitle={Proceedings of the AAAI conference on artificial intelligence},
  volume={38},
  number={5},
  pages={4296--4304},
  year={2024}
}

@article{song2020score,
  title={Score-based generative modeling through stochastic differential equations},
  author={Song, Yang and Sohl-Dickstein, Jascha and Kingma, Diederik P and Kumar, Abhishek and Ermon, Stefano and Poole, Ben},
  journal={arXiv preprint arXiv:2011.13456},
  year={2020}
}

@article{ddpm,
  title={Denoising diffusion probabilistic models},
  author={Ho, Jonathan and Jain, Ajay and Abbeel, Pieter},
  journal={Advances in neural information processing systems},
  volume={33},
  pages={6840--6851},
  year={2020}
}

@article{ddim,
  title={Denoising diffusion implicit models},
  author={Song, Jiaming and Meng, Chenlin and Ermon, Stefano},
  journal={arXiv preprint arXiv:2010.02502},
  year={2020}
}

@article{peng2024dreambench++,
  title={Dreambench++: A human-aligned benchmark for personalized image generation},
  author={Peng, Yuang and Cui, Yuxin and Tang, Haomiao and Qi, Zekun and Dong, Runpei and Bai, Jing and Han, Chunrui and Ge, Zheng and Zhang, Xiangyu and Xia, Shu-Tao},
  journal={arXiv preprint arXiv:2406.16855},
  year={2024}
}

@article{oquab2023dinov2,
  title={Dinov2: Learning robust visual features without supervision},
  author={Oquab, Maxime and Darcet, Timoth{\'e}e and Moutakanni, Th{\'e}o and Vo, Huy and Szafraniec, Marc and Khalidov, Vasil and Fernandez, Pierre and Haziza, Daniel and Massa, Francisco and El-Nouby, Alaaeldin and others},
  journal={arXiv preprint arXiv:2304.07193},
  year={2023}
}

@article{xiao2024fastcomposer,
  title={Fastcomposer: Tuning-free multi-subject image generation with localized attention},
  author={Xiao, Guangxuan and Yin, Tianwei and Freeman, William T and Durand, Fr{\'e}do and Han, Song},
  journal={International Journal of Computer Vision},
  pages={1--20},
  year={2024},
  publisher={Springer}
}

@inproceedings{hyperdreambooth,
  title={Hyperdreambooth: Hypernetworks for fast personalization of text-to-image models},
  author={Ruiz, Nataniel and Li, Yuanzhen and Jampani, Varun and Wei, Wei and Hou, Tingbo and Pritch, Yael and Wadhwa, Neal and Rubinstein, Michael and Aberman, Kfir},
  booktitle={Proceedings of the IEEE/CVF conference on computer vision and pattern recognition},
  pages={6527--6536},
  year={2024}
}

@article{hu2022lora,
  title={Lora: Low-rank adaptation of large language models.},
  author={Hu, Edward J and Shen, Yelong and Wallis, Phillip and Allen-Zhu, Zeyuan and Li, Yuanzhi and Wang, Shean and Wang, Lu and Chen, Weizhu and others},
  journal={ICLR},
  volume={1},
  number={2},
  pages={3},
  year={2022}
}

@article{disenbooth,
  title={Disenbooth: Identity-preserving disentangled tuning for subject-driven text-to-image generation},
  author={Chen, Hong and Zhang, Yipeng and Wu, Simin and Wang, Xin and Duan, Xuguang and Zhou, Yuwei and Zhu, Wenwu},
  journal={arXiv preprint arXiv:2305.03374},
  year={2023}
}

@inproceedings{textual_inversion,
  title={An Image is Worth One Word: Personalizing Text-to-Image Generation using Textual Inversion},
  author={Gal, Rinon and Alaluf, Yuval and Atzmon, Yuval and Patashnik, Or and Bermano, Amit Haim and Chechik, Gal and Cohen-or, Daniel},
  booktitle={The Eleventh International Conference on Learning Representations},
year={2022}
}

@inproceedings{dreambooth,
  title={Dreambooth: Fine tuning text-to-image diffusion models for subject-driven generation},
  author={Ruiz, Nataniel and Li, Yuanzhen and Jampani, Varun and Pritch, Yael and Rubinstein, Michael and Aberman, Kfir},
  booktitle={Proceedings of the IEEE/CVF conference on computer vision and pattern recognition},
  pages={22500--22510},
  year={2023}
}

@inproceedings{custom_diffusion,
  title={Multi-concept customization of text-to-image diffusion},
  author={Kumari, Nupur and Zhang, Bingliang and Zhang, Richard and Shechtman, Eli and Zhu, Jun-Yan},
  booktitle={Proceedings of the IEEE/CVF conference on computer vision and pattern recognition},
  pages={1931--1941},
  year={2023}
}

@article{ipadapter,
  title={Ip-adapter: Text compatible image prompt adapter for text-to-image diffusion models},
  author={Ye, Hu and Zhang, Jun and Liu, Sibo and Han, Xiao and Yang, Wei},
  journal={arXiv preprint arXiv:2308.06721},
  year={2023}
}

@inproceedings{wei2023elite,
  title={Elite: Encoding visual concepts into textual embeddings for customized text-to-image generation},
  author={Wei, Yuxiang and Zhang, Yabo and Ji, Zhilong and Bai, Jinfeng and Zhang, Lei and Zuo, Wangmeng},
  booktitle={Proceedings of the IEEE/CVF International Conference on Computer Vision},
  pages={15943--15953},
  year={2023}
}

@inproceedings{ldm,
  title={High-resolution image synthesis with latent diffusion models},
  author={Rombach, Robin and Blattmann, Andreas and Lorenz, Dominik and Esser, Patrick and Ommer, Bj{\"o}rn},
  booktitle={Proceedings of the IEEE/CVF conference on computer vision and pattern recognition},
  pages={10684--10695},
  year={2022}
}

@article{imagen,
  title={Photorealistic text-to-image diffusion models with deep language understanding},
  author={Saharia, Chitwan and Chan, William and Saxena, Saurabh and Li, Lala and Whang, Jay and Denton, Emily L and Ghasemipour, Kamyar and Gontijo Lopes, Raphael and Karagol Ayan, Burcu and Salimans, Tim and others},
  journal={Advances in neural information processing systems},
  volume={35},
  pages={36479--36494},
  year={2022}
}

@article{dalle2,
  title={Hierarchical text-conditional image generation with clip latents},
  author={Ramesh, Aditya and Dhariwal, Prafulla and Nichol, Alex and Chu, Casey and Chen, Mark},
  journal={arXiv preprint arXiv:2204.06125},
  volume={1},
  number={2},
  pages={3},
  year={2022}
}

@inproceedings{cfg,
  title={Classifier-Free Diffusion Guidance},
  author={Ho, Jonathan and Salimans, Tim},
  booktitle={NeurIPS 2021 Workshop on Deep Generative Models and Downstream Applications},
year={2022}
}

@inproceedings{clip,
  title={Learning transferable visual models from natural language supervision},
  author={Radford, Alec and Kim, Jong Wook and Hallacy, Chris and Ramesh, Aditya and Goh, Gabriel and Agarwal, Sandhini and Sastry, Girish and Askell, Amanda and Mishkin, Pamela and Clark, Jack and others},
  booktitle={International conference on machine learning},
  pages={8748--8763},
  year={2021},
  organization={PmLR}
}

@inproceedings{disenvisioner,
  title={DisEnvisioner: Disentangled and Enriched Visual Prompt for Customized Image Generation},
  author={He, Jing and Haodong, LI and Shen, Guibao and Yingjie, CAI and Qiu, Weichao and Chen, Ying-Cong and others},
  booktitle={The Thirteenth International Conference on Learning Representations},
  year={2024}
}

@inproceedings{han2023svdiff,
  title={Svdiff: Compact parameter space for diffusion fine-tuning},
  author={Han, Ligong and Li, Yinxiao and Zhang, Han and Milanfar, Peyman and Metaxas, Dimitris and Yang, Feng},
  booktitle={Proceedings of the IEEE/CVF International Conference on Computer Vision},
  pages={7323--7334},
  year={2023}
}

@misc{vae,
  title={Auto-encoding variational bayes},
  author={Kingma, Diederik P and Welling, Max and others},
  year={2013},
  publisher={Banff, Canada}
}

@misc{Freepik,
  title        = {Freepik},
  howpublished = {\url{https://www.freepik.com/}},   
  year         = {2025},                     
}

@inproceedings{blora,
  title={Implicit style-content separation using b-lora},
  author={Frenkel, Yarden and Vinker, Yael and Shamir, Ariel and Cohen-Or, Daniel},
  booktitle={European Conference on Computer Vision},
  pages={181--198},
  year={2024},
  organization={Springer}
}

@inproceedings{paircustomization,
  title={Customizing text-to-image models with a single image pair},
  author={Jones, Maxwell and Wang, Sheng-Yu and Kumari, Nupur and Bau, David and Zhu, Jun-Yan},
  booktitle={SIGGRAPH Asia 2024 Conference Papers},
  pages={1--13},
  year={2024}
}

@inproceedings{ziplora,
  title={Ziplora: Any subject in any style by effectively merging loras},
  author={Shah, Viraj and Ruiz, Nataniel and Cole, Forrester and Lu, Erika and Lazebnik, Svetlana and Li, Yuanzhen and Jampani, Varun},
  booktitle={European Conference on Computer Vision},
  pages={422--438},
  year={2024},
  organization={Springer}
}

@inproceedings{syncd,
  title={Generating multi-image synthetic data for text-to-image customization},
  author={Kumari, Nupur and Yin, Xi and Zhu, Jun-Yan and Misra, Ishan and Azadi, Samaneh},
  booktitle={Proceedings of the IEEE/CVF International Conference on Computer Vision},
  pages={16524--16534},
  year={2025}
}

@inproceedings{ominicontrol,
  title={Ominicontrol: Minimal and universal control for diffusion transformer},
  author={Tan, Zhenxiong and Liu, Songhua and Yang, Xingyi and Xue, Qiaochu and Wang, Xinchao},
  booktitle={Proceedings of the IEEE/CVF International Conference on Computer Vision},
  pages={14940--14950},
  year={2025}
}

@article{tokenverse,
  title={Tokenverse: Versatile multi-concept personalization in token modulation space},
  author={Garibi, Daniel and Yadin, Shahar and Paiss, Roni and Tov, Omer and Zada, Shiran and Ephrat, Ariel and Michaeli, Tomer and Mosseri, Inbar and Dekel, Tali},
  journal={ACM Transactions On Graphics (TOG)},
  volume={44},
  number={4},
  pages={1--11},
  year={2025},
  publisher={ACM New York, NY, USA}
}

@inproceedings{ssrencoder,
  title={Ssr-encoder: Encoding selective subject representation for subject-driven generation},
  author={Zhang, Yuxuan and Song, Yiren and Liu, Jiaming and Wang, Rui and Yu, Jinpeng and Tang, Hao and Li, Huaxia and Tang, Xu and Hu, Yao and Pan, Han and others},
  booktitle={Proceedings of the IEEE/CVF Conference on Computer Vision and Pattern Recognition},
  pages={8069--8078},
  year={2024}
}

@inproceedings{dit,
  title={Scalable diffusion models with transformers},
  author={Peebles, William and Xie, Saining},
  booktitle={Proceedings of the IEEE/CVF international conference on computer vision},
  pages={4195--4205},
  year={2023}
}
}

\clearpage
\setcounter{page}{1}
\maketitlesupplementary

\newcommand{\bx}{\mathbf{x}}
\newcommand{\bz}{\mathbf{z}}
\newcommand{\bt}{\mathbf{t}}
\newcommand{\bepsilon}{\boldsymbol{\epsilon}}
\newcommand{\bzero}{\mathbf{0}}
\newcommand{\bI}{\mathbf{I}}

\section{Derivation of Equation~\ref{eq:excl_loss}}
\label{sec:appendix_derivation}

The objective of our exclusion loss is to ensure that the residual token $\mathbf{t}_{\text{residual}}^{(i)}$ remains uninformative regarding the target concept when reconstructing other images $\mathbf{x}^{(j)}$ where $j \neq i$. As explained in Section~\ref{sec:3.2}, this is achieved by minimizing the KL divergence between the distribution conditioned on the residual token and the unconditional distribution:

\vspace{-4mm}
\begin{equation}
    \begin{split}
        \mathcal{L}_{\text{KL}} = \mathbb{E}_{i, \mathbf{z}_t^{(j\neq i)}, t} \left[ D_{\text{KL}} \left( p_{\theta}(\mathbf{z}_{t-1}^{(j)} | \mathbf{z}_{t}^{(j)}, t, \varnothing) \right. \right. \\ \left. \left. \parallel p_{\theta}(\mathbf{z}_{t-1}^{(j)} | \mathbf{z}_{t}^{(j)}, t, \Gamma(c_{\text{residual}}^{(i)})) \right) \right].
    \end{split}
\end{equation}
\vspace{-2mm}

In diffusion models~\cite{ldm}, the reverse process $p_{\theta}(\mathbf{z}_{t-1} | \mathbf{z}_{t}, t, \mathbf{c})$ is parameterized as a Gaussian distribution $\mathcal{N}(\mathbf{z}_{t-1}; \boldsymbol{\mu}_{\theta}(\mathbf{z}_{t}, t, \mathbf{c}), \boldsymbol{\Sigma}_{\theta}(\mathbf{z}_{t}, t))$. Following standard practice~\cite{ddpm}, we fix the covariance matrix $\boldsymbol{\Sigma}_{\theta}$ to a time-dependent constant $\beta_t \mathbf{I}$. The cumulative signal scale $\bar{\alpha}_t$ is defined as $\bar{\alpha}_t = \prod_{s=1}^{t} (1 - \beta_s)$ following DDPM notation~\cite{ddpm}, based on the noise schedule $\beta_t$. Under the assumption of equal covariance, the KL divergence between two Gaussians is proportional to the squared Euclidean distance between their means:
\vspace{-2mm}

\begin{equation}
    D_{\text{KL}}(\mathcal{N}(\boldsymbol{\mu}_1, \boldsymbol{\Sigma}) \parallel \mathcal{N}(\boldsymbol{\mu}_2, \boldsymbol{\Sigma})) \propto \left\|\boldsymbol{\mu}_1 - \boldsymbol{\mu}_2\right\|_2^2.
\end{equation}
\vspace{-2mm}

Thus, minimizing $\mathcal{L}_{\text{KL}}$ is equivalent to minimizing the distance between the predicted means given the null condition $\varnothing$ and the residual condition $c_{\text{residual}}^{(i)}$. The mean $\boldsymbol{\mu}_{\theta}$ is related to the noise prediction $\boldsymbol{\epsilon}_{\theta}$ by:

\vspace{-2mm}

\begin{equation}
    \boldsymbol{\mu}_{\theta}(\mathbf{z}_{t}, t, \mathbf{c}) = \frac{1}{\sqrt{1-\beta_t}} \left( \mathbf{z}_t - \frac{\beta_t}{\sqrt{1-\bar{\alpha}_t}} \boldsymbol{\epsilon}_{\theta}(\mathbf{z}_t, t, \mathbf{c}) \right). 
\end{equation}
\vspace{-2mm}

Substituting this into the distance metric, the terms involving $\mathbf{z}_t$ cancel out, simplifying the objective to the difference between noise predictions:

\begin{align} 
\label{eq:mean_diff} 
    & \quad \left\|\boldsymbol{\mu}_{\theta}(\mathbf{z}_{t}^{(j)}, t, \varnothing) - \boldsymbol{\mu}_{\theta}(\mathbf{z}_{t}^{(j)}, t, \Gamma(c_{\text{residual}}^{(i)}))\right\|_2^2 \\ 
    &= \left\| \frac{1}{\sqrt{1-\beta_t}} \left( \mathbf{z}_t^{(j)} - \frac{\beta_t}{\sqrt{1-\bar{\alpha}_t}} \boldsymbol{\epsilon}_{\theta}(\mathbf{z}_t^{(j)}, t, \varnothing) \right) \right. \nonumber \\ 
    &\quad \left. - \frac{1}{\sqrt{1-\beta_t}} \left( \mathbf{z}_t^{(j)} - \frac{\beta_t}{\sqrt{1-\bar{\alpha}_t}} \boldsymbol{\epsilon}_{\theta}(\mathbf{z}_t^{(j)}, t, \Gamma(c_{\text{residual}}^{(i)})) \right) \right\|_2^2 \nonumber \\ 
    &= \frac{\beta_t^2}{(1-\bar{\alpha}_t) (1-\beta_t)} \left\| \boldsymbol{\epsilon}_{\theta}(\mathbf{z}_t^{(j)}, t, \Gamma(c_{\text{residual}}^{(i)})) - \boldsymbol{\epsilon}_{\theta}(\mathbf{z}_t^{(j)}, t, \varnothing) \right\|_2^2. \nonumber 
\end{align}

By absorbing the scalar coefficients $\frac{\beta_t^2}{(1-\bar{\alpha}_t) (1-\beta_t)}$ into the weighting term $w_t$ and summing over all valid pairs $(i, j)$ where $j \neq i$, we arrive at the final exclusion loss presented in Equation~\ref{eq:excl_loss}.

\section{More Experimental Details}

\paragraph{Algorithms.}

The exact implementation of ConceptPrism are summarized in Algorithm~\ref{alg:stage1} and Algorithm~\ref{alg:stage2}. Algorithm~\ref{alg:stage1} outlines the token optimization stage in Section~\ref{sec:3.1} to \ref{sec:3.3}, where the target and residual tokens are jointly optimized. Algorithm~\ref{alg:stage2} describes the subsequent concept disentangled fine-tuning stage in Section~\ref{sec:3.4}. For computational efficiency, we approximate the exclusion loss $\mathcal{L}_{\text{excl}}$ during the token optimization. Instead of utilizing the entire set of $N-1$ negative samples, we randomly sample a subset of 3 indices $\mathcal{J} \subset \{k \mid k \neq i\}$ at each iteration.

\paragraph{Image Captions on Section~\ref{sec:3.3}.}

We employed a Vision Language Model (VLM), specifically Gemini 2.5 Flash, to generate descriptive captions for each reference image. These captions serve as the initialization for residual tokens (Section~\ref{sec:3.3}). The specific prompt used is provided below. Note that this initialization captures the overall scene, ensuring the initial tokens are not biased toward either the target concept or specific residuals.

\begin{tcolorbox}[title=Prompt for Scene Description]
You are an AI that describes visual scenes in detail. Your mission is to describe a given image using 8-32 words of text.

\subsection*{Instructions}
\begin{enumerate}
    \item \textbf{Purpose of Description:} Describe the image in sufficient detail that the original scene can be visually \textbf{reconstructed} based \textbf{solely on the text description}.
    \item \textbf{Elements to Describe:} You must include not only the main \textbf{subject} but also the core visual elements that constitute the scene, such as its key \textbf{attributes, background, composition, lighting, and style}.
    \item All descriptions must be written in English and be between 8 and 32 words.
\end{enumerate}
\end{tcolorbox}

\begin{algorithm}[h]
\caption{Token Optimization}
\label{alg:stage1}
\begin{algorithmic}[1]
\Require Reference images $\mathcal{X} = \{\bx^{(i)}\}_{i=1}^N$, Pre-trained Model $\epsilon_\theta$, Exclusion weight $\beta$, Total timesteps $T$
\Ensure Optimized tokens $\bt_{\text{target}}, \{\bt_{\text{residual}}^{(i)}\}_{i=1}^N$

\State Initialize $\bt_{\text{target}}$ randomly
\State Initialize $\{\bt_{\text{residual}}^{(i)}\}_{i=1}^N$ with caption embeddings of $\{\bx^{(i)}\}_{i=1}^N$
\While{not converged}
    \State Sample $i \sim \mathcal{U}(1, N)$ and $\bx^{(i)} \sim \mathcal{X}$
    \State Sample $\bepsilon \sim \mathcal{N}(\mathbf{0}, \mathbf{I}), t \sim \mathcal{U}(\{1,\dots T\})$
    \State \textcolor{gray}{// Compute Reconstruction Loss}
    \State $c^{(i)} \leftarrow ``\bt_{\text{target}} \text{ with } \bt_{\text{residual}}^{(i)}"$
    \State $\mathcal{L}_{\text{rec}} \leftarrow \| \bepsilon - \epsilon_\theta(\bz_t^{(i)}, t, \Gamma(c^{(i)})) \|_2^2$
    
    \State \textcolor{gray}{// Compute Exclusion Loss}
    \State Sample $\mathcal{J} \subset \{k \mid k \neq i\}$ s.t. $|\mathcal{J}| = 3$
    \State $\mathcal{L}_{\text{excl}} \leftarrow 0$
    \For{$j \in \mathcal{J}$}
        \State $\mathcal{L}_{\text{excl}} \leftarrow \mathcal{L}_{\text{excl}} + \| \epsilon_\theta(\bz_t^{(j)}, t, \Gamma(\bt_{\text{residual}}^{(i)})) - \epsilon_\theta(\bz_t^{(j)}, t, \varnothing) \|_2^2$
    \EndFor
    
    \State Update $\bt_{\text{target}}, \bt_{\text{residual}}^{(i)}$ via $\nabla (\mathcal{L}_{\text{rec}} + \beta \mathcal{L}_{\text{excl}})$
\EndWhile
\end{algorithmic}
\end{algorithm}
\vspace{-2mm}

\begin{algorithm}[h]
\caption{Concept Disentangled Fine-Tuning}
\label{alg:stage2}
\begin{algorithmic}[1]
\Require Reference images $\mathcal{X} = \{\bx^{(i)}\}_{i=1}^{N}$, Optimized tokens $\bt_{\text{target}}, \{\bt_{\text{residual}}^{(i)}\}_{i=1}^{N}$, Model $\epsilon_\theta$, Total timesteps $T$
\Ensure Personalized Model $\epsilon_{\theta^*}$

\State Fix $\bt_{\text{target}}$ and $\{\bt_{\text{residual}}^{(i)}\}_{i=1}^N$
\State Inject LoRA parameters $\theta_{\text{lora}}$ into attention layers of $\epsilon_\theta$
\While{not converged}
    \State Sample batch $(i, \bx^{(i)})$ from $\mathcal{X}$
    \State Sample $\bepsilon \sim \mathcal{N}(\mathbf{0}, \mathbf{I}), t \sim \mathcal{U}(\{1,\dots T\})$
    \State \textcolor{gray}{// Fine-tune with disentangled tokens}
    \State $c \leftarrow ``\bt_{\text{target}} \text{ with } \bt_{\text{residual}}^{(i)}"$
    \State $\mathcal{L}_{\text{ft}} \leftarrow \| \bepsilon - \epsilon_\theta(\bz_t^{(i)}, t, \Gamma(c)) \|_2^2$
    \State Update $\theta_{\text{lora}}$ via $\nabla \mathcal{L}_{\text{ft}}$
\EndWhile
\end{algorithmic}
\end{algorithm}

\vspace{-6mm}
\section{More Ablation Studies}

\paragraph{Impact of Target Token Length.}

We investigated the impact of the target token length ($n$) on performance. A longer token sequence theoretically provides greater capacity to encode visual information. We evaluated lengths of $n \in \{1, 2, 4, 8\}$ as summarized in Table~\ref{tab:target_token_length}. The results indicate that $n=1$ yields the optimal balance. Increasing $n$ marginally improves Concept Fidelity (DINO) but significantly degrades Text Alignment (CLIP-T).

This trade-off suggests that a high-capacity target token captures excessive details and leads to conflicts with the text prompt. This phenomenon may stem from the transformer, which struggles to handle extended contexts, or the concept entanglement of the target token. Specifically, an over-parameterized target token risks memorizing the entire reference images including residuals, regardless of the exclusion loss applied to residual tokens. While introducing a mechanism to minimize the information overlap between target and residual tokens remains a promising direction for future work, our empirical results confirm that a single token ($n=1$) provides sufficient capacity for effectively capturing most concepts.

\vspace{-2mm}
\begin{table}[h]
  \centering
    \begin{tabular}{ccccc} 
      \toprule
      $n$ & 1 & 2 & 4 & 8 \\
      \midrule
      CLIP-T & 0.357 & 0.352 & 0.339 & 0.325 \\
      DINO   & 0.210 & 0.207 & 0.212 & 0.220 \\
      \bottomrule
    \end{tabular}
  \caption{\textbf{Impact of Target Token Length.} $n$ denotes the length of the target token. We adopted $n=1$ for our main experiments.}
  \label{tab:target_token_length}
\end{table}

\vspace{-6mm}

\paragraph{Impact of Exclusion Loss Weight.}

We analyzed the impact of the exclusion loss weight $\beta$ on model performance. (Table~\ref{tab:ablation_beta}). The results show that introducing $\mathcal{L}_{excl}$ significantly improves Concept Fidelity (DINO) compared to the baseline without exclusion loss ($\beta=0$). This confirms that our exclusion objective effectively prevents residual tokens from absorbing the target concept.

However, an excessively large $\beta$ leads to a degradation in Text Alignment (CLIP-T). A strong penalty forces residual tokens to discard necessary image-specific details and converge towards a null condition. Consequently, the target token inadvertently captures this residual information to minimize the reconstruction loss. Our experiments suggest that $\beta=0.05$ yields the optimal trade-off.

\begin{table}[h]
  \centering
    \begin{tabular}{cccccc} 
      \toprule
      $\beta$ & 0 & 0.01 & 0.05 & 0.1 & 0.5 \\
      \midrule
      CLIP-T & 0.358 & 0.356 & 0.354 & 0.346 & 0.347 \\
      DINO   & 0.183 & 0.198 & 0.207 & 0.209 & 0.206 \\
      \bottomrule
    \end{tabular}
  \caption{\textbf{Impact of Exclusion Loss Weight.} We adopted the exclusion loss weight $\beta=0.05$ for our main experiments.}
  \label{tab:ablation_beta}
\end{table}

\vspace{-4mm}
\section{More Comparisons with Prior Works}

\paragraph{VLM Evaluations.}

\begin{figure*}
  \centering
  \includegraphics[width=1.0\textwidth]{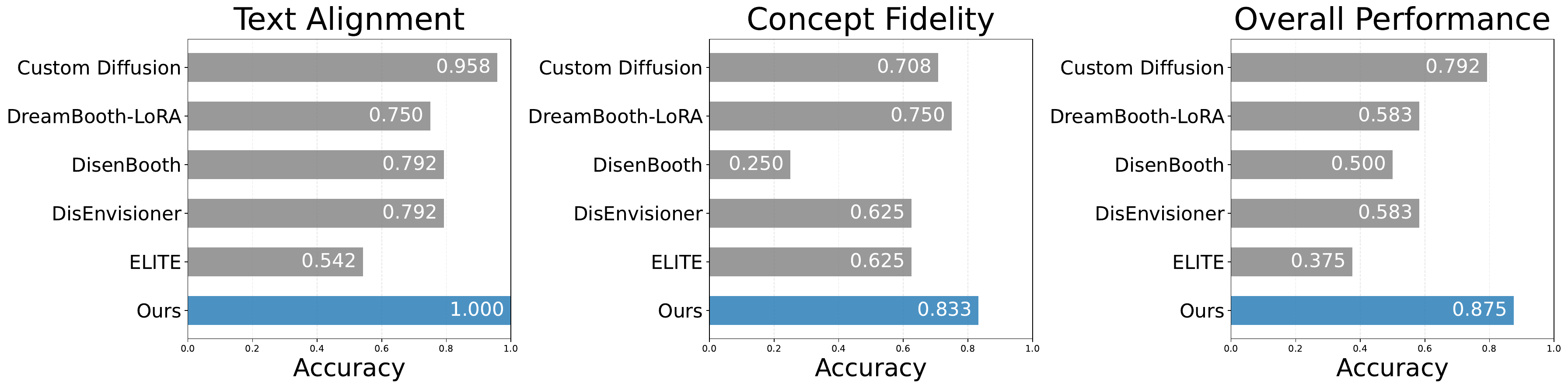}
  \caption{\textbf{VLM Evaluations.} We prompted GPT-4o to assess generated images on three criteria: Concept Fidelity, Text Alignment, and Overall Preference. The charts display the satisfaction rates for each method. Our method consistently outperforms baselines across all metrics, validating the effectiveness of our disentanglement framework.}
  \label{fig:vlm_results}
\end{figure*}

Standard quantitative metrics often diverge from human perception. To assess practical performance, we conducted an evaluation using a VLM. Specifically, we employed GPT-4o on 24 random concept-prompt pairs from DreamBench. The VLM evaluated images based on three criteria: (1) \textit{Concept Fidelity}, (2) \textit{Text Alignment}, and (3) \textit{Overall Preference}, providing a binary decision (satisfy/unsatisfy) for each. Figure~\ref{fig:vlm_results} summarizes the satisfaction rates. Our method consistently achieved the highest scores across all criteria, demonstrating the effectiveness of our disentanglement framework.

The specific prompts used for the evaluation are detailed below.

\vspace{1em}
\noindent
\textbf{Evaluation Prompts}

\begin{itemize}[leftmargin=*]

    \item \textbf{Text Alignment:}\begin{quote}\itshape``You are an impartial and unambiguous judge who must evaluate the quality of the given picture. You must assign either 1 (satisfies the condition) or 0 (does not). The condition is: \\
    `Does the picture reflect the following text prompt? - Text Prompt: \{text\_prompt\}'
    
    Please output a single digit (1 or 0).''
    \end{quote}

    \item \textbf{Concept Fidelity:}
    \begin{quote}
    \itshape
    ``You are an impartial and unambiguous judge who must evaluate the quality of the given picture. You must assign either 1 (satisfies the condition) or 0 (does not). The condition is: \\
    `Does the picture faithfully follow the visual details of the source images?'
    
    Please output a single digit (1 or 0).''
    \end{quote}

    \item \textbf{Overall Preference:}
    \begin{quote}
    \itshape
    ``You are an impartial and unambiguous judge who must evaluate the quality of the given picture. You must assign either 1 (satisfies the condition) or 0 (does not). The condition is: \\
    `Considering both the text prompt and the source images, does the given picture follow both the text prompt and source visual details?'
    
    Please output a single digit (1 or 0).''
    \end{quote}
\end{itemize}


\paragraph{Comparisons with Recent Methods on DiT Backbone.}
\begin{figure*}[t]
  \centering
  \includegraphics[width=1.0\textwidth]{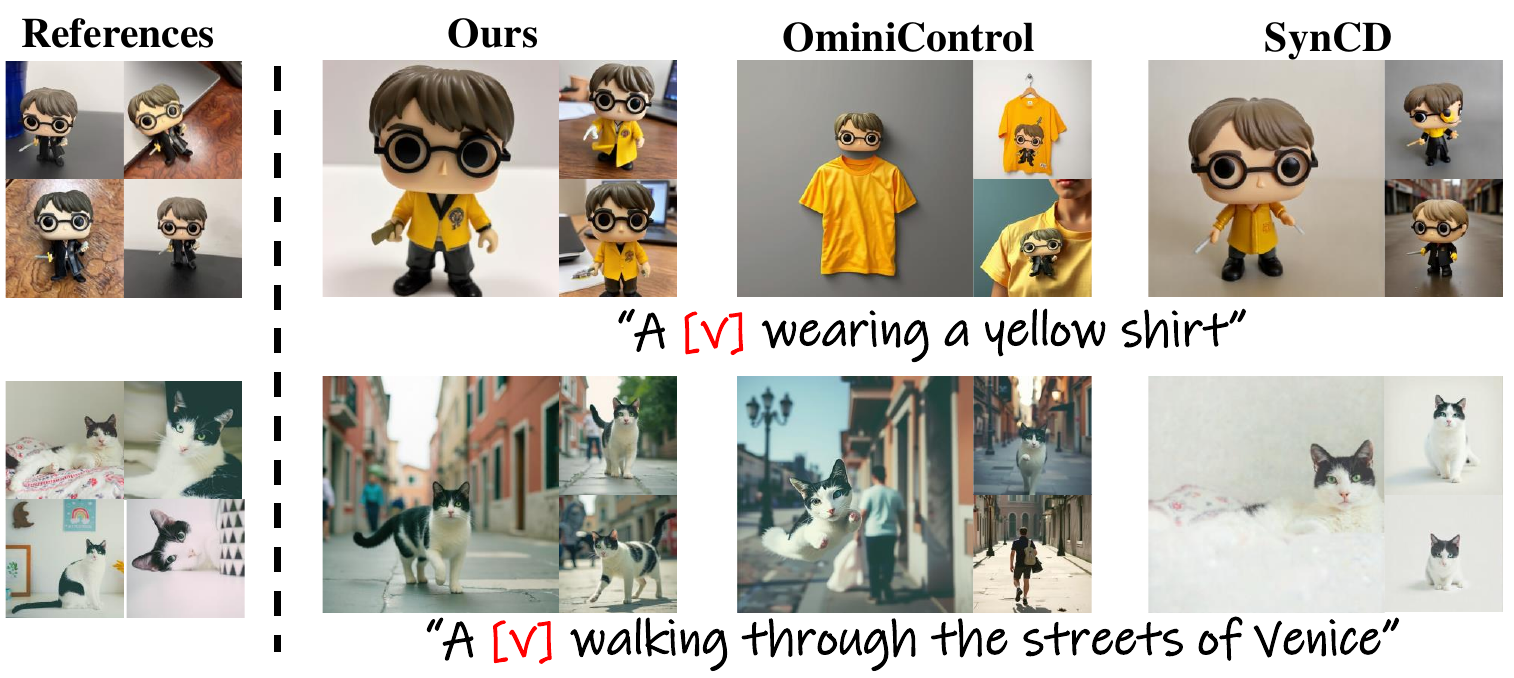}
  \caption{\textbf{Comparison with recent methods using a DiT backbone.} Being architecture-agnostic, ConceptPrism is effectively applicable to the latest DiT-based generative models. Our disentanglement strategy enables a deep understanding of the target concept, excelling in handling prompts that require attribute variations of the target.}
  \vspace{-4mm}
  \label{fig:flux_qualitative}
\end{figure*}

\begin{table}[h]
  \centering
  \caption{\textbf{Quantitative Evaluations on DiT Backbone.} ConceptPrism achieves balanced, high performance compared to personalization techniques specialized for DiT architectures, which demonstrates the practical utility of our method.}
    \begin{tabular}{lcc}
        \toprule
      Method & CLIP-T $\uparrow$ & DINO $\uparrow$ \\
      \midrule
       SynCD & 0.300 & \textbf{0.269} \\
       OminiCtrl & \underline{0.346} & 0.174 \\
       Ours & \textbf{0.362} & \underline{0.213} \\
      \bottomrule
    \end{tabular}
  \label{tab:flux_quant}
\end{table}

\begin{figure*}
  \centering
  \includegraphics[width=1.0\textwidth]{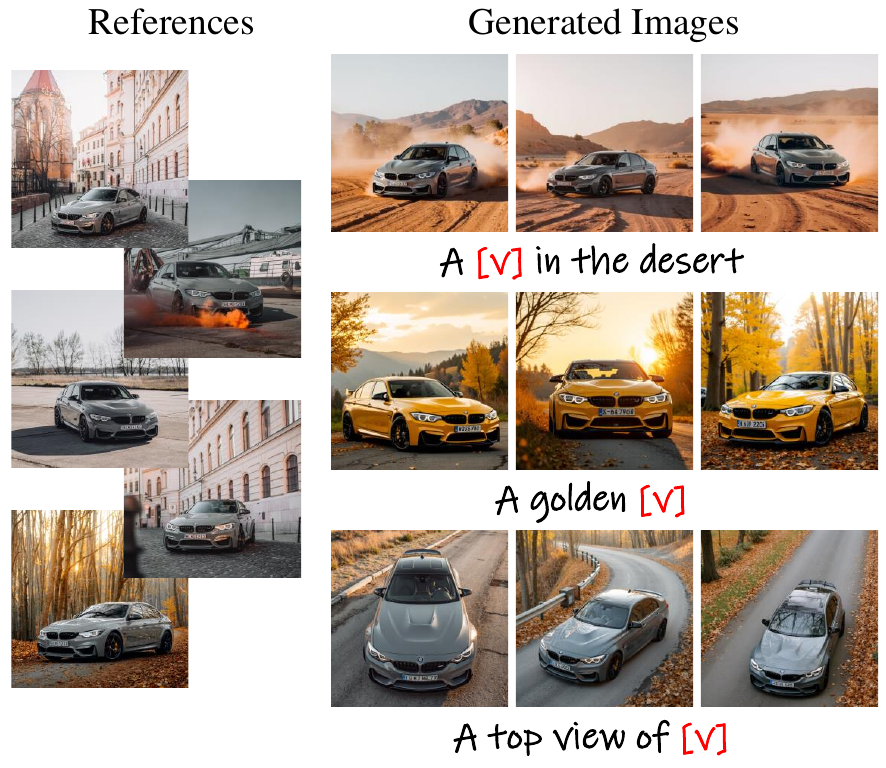}
  \caption{\textbf{Application on FLUX.1.} Since our framework operates on the text embedding space, it is agnostic to the backbone architecture of the diffusion model. We demonstrate this broad applicability by successfully generating personalized images using the FLUX.1-dev model.}
  \label{fig:additional_flux}
\end{figure*}

Recent text-to-image generative models rely on Diffusion Transformer (DiT) architectures to achieve high performance~\cite{syncd,ominicontrol}. ConceptPrism is universally applicable to various generative model architectures as it focuses on token optimization without requiring additional networks. Figure~\ref{fig:flux_qualitative} and Table~\ref{tab:flux_quant} compare our method with personalization techniques specialized for DiT structures using a 4-bit quantized FLUX.1 model. Our method outperforms these specialized baselines by achieving balanced, high performance across concept fidelity and text alignment. Such results demonstrate our superior disentanglement capabilities in capturing shared concepts regardless of the backbone architecture. Further examples of generations on the DiT backbone are presented in Figure~\ref{fig:additional_flux}.


\vspace{6mm}

\paragraph{Comparisons with Encoder-based Methods.}
\begin{table}[h]
  \centering
  \caption{\textbf{Time Comparison with Encoder-based Methods.} We measure the training and inference time for personalized T2I generation using a 4-bit quantized FLUX.1 model. Our method incurs overhead compared to encoder-based methods due to the requirement of concept-wise training. However, it enables lighter inference by removing the need for reference information after training is complete.}
    \begin{tabular}{lccc} 
      \toprule
    Method & Inference Time (s) & Training Time (s) \\
    \hline
    Ours & 18.186 & 507.431 \\
    SynCD & 38.366 & - \\
    OminiCtrl & 21.093 & - \\
    \bottomrule
    \end{tabular}
  \label{tab:time_vs_encoder_based}
\end{table}

Comparisons with encoder-based personalization methods~\cite{syncd,ominicontrol} further verify the practical efficiency of ConceptPrism. Optimization-based approaches, including our method, incur overhead by requiring concept-wise training. As shown in Table~\ref{tab:time_vs_encoder_based}, our method takes approximately 8 minutes to train a personalized concept on a 4-bit quantized FLUX.1 model using a single H100 GPU. This duration remains within a practical range for applications involving repeated inference. In return, our method eliminates the need for reference images during the inference stage. Encoder-based methods must inject reference information into the input context for every generation, leading to high computational costs. Furthermore, these methods often suffer from performance degradation as the context length increases with multiple reference images.
\section{Additional Experiments}

\begin{figure*}[t]
  \centering
  \includegraphics[width=1.0\textwidth]{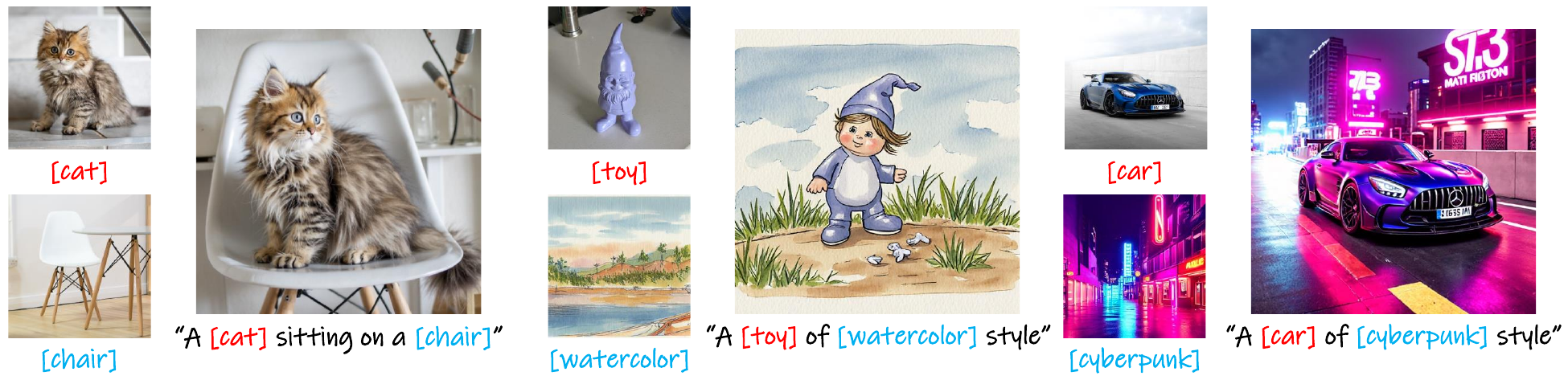}
  \caption{\textbf{Multi Concept Composition.} Each personalized concept is learned independently following our protocol and merged via LoRA merge to generate multiple concepts simultaneously. Results are based on the FLUX.1 backbone model.}
  \label{fig:multi_concept}
\end{figure*}

\begin{figure*}
  \centering
  \includegraphics[width=0.8\textwidth]{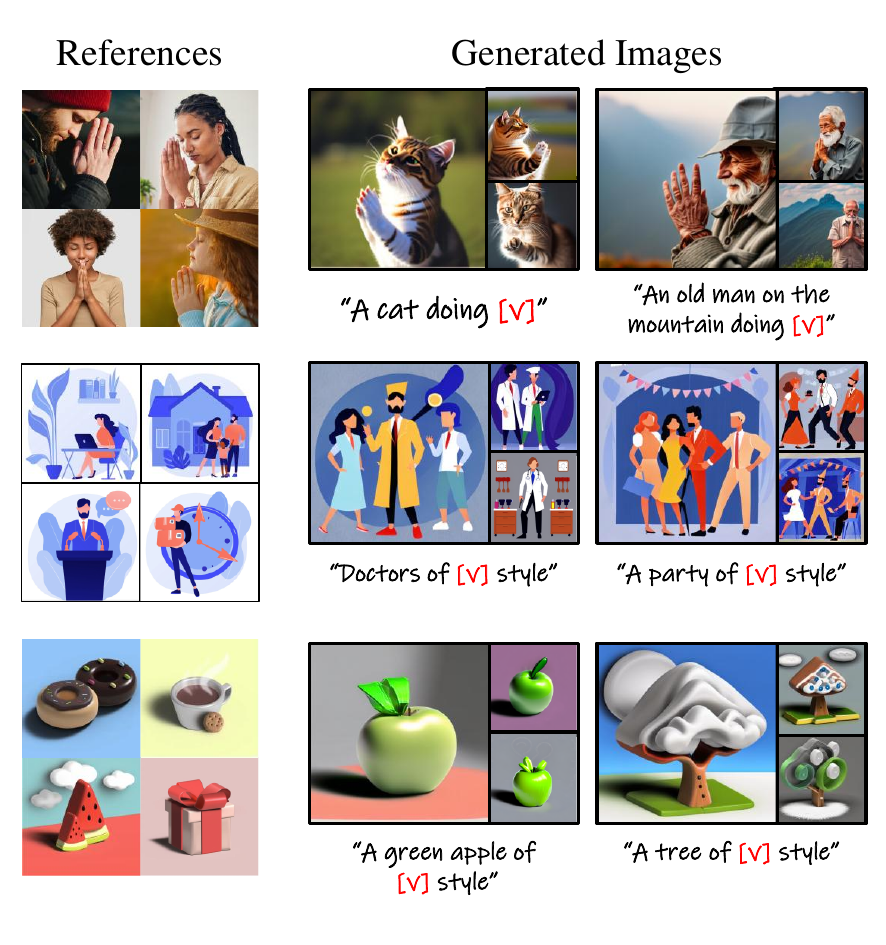}
  \vspace{-6mm}
  \caption{\textbf{Additional results on abstract concepts.} Our method effectively disentangles and learns abstract concepts from various datasets, including styles and actions.}
  \label{fig:additional_abstract_concept}
\end{figure*}
\begin{figure*}
  \centering
  \includegraphics[width=1.0\textwidth]{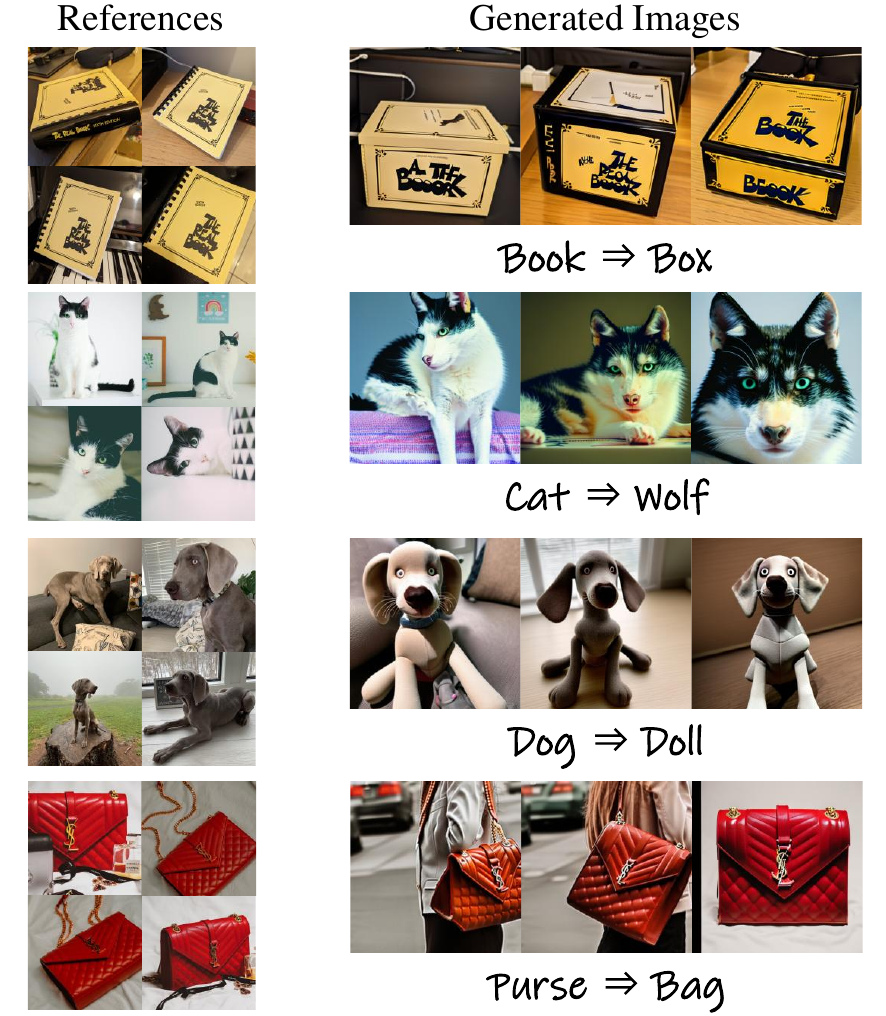}
  \caption{\textbf{Flexible modifications on learned concepts.} Since our method learns without class noun guidance, the learned token can be flexibly applied to different subject classes. To generate these images, we used prompts specifying a new subject class, such as ``A photo of [target] box'' or ``A photo of [target] wolf''.}
  \label{fig:additional_class_noun}
\end{figure*}


\paragraph{Multi-concept Composition.}

Our method is effectively applicable to multi-concept composition. We learn each personalized concept independently according to our protocol. These concepts are then combined through LoRA merge, requiring no joint training. Our residual tokens absorb irrelevant information, enabling target tokens to retain only essential features. Such well-disentangled tokens can be merged without interference. Including these target tokens in the text condition allows for the composition of multiple concepts in diverse forms. Figure~\ref{fig:multi_concept} illustrates successful compositions between subjects and styles, utilizing FLUX.1 as the backbone model.

\paragraph{More Flexible Modifications on Learned Concepts.}

Most personalization methods bind the learned concept to a specific coarse class noun (e.g., ``dog'' or ``book'') during the optimization process. In contrast, ConceptPrism extracts the target concept without such semantic priors. This independence allows for fundamental semantic shifts during generation. As shown in Figure~\ref{fig:additional_class_noun}, our model can transfer learned visual attributes onto entirely different object categories (e.g., transforming a specific book design into a box). This capability demonstrates that our token captures intrinsic visual structures rather than being tied to a linguistic class. We conducted these experiments using the CustomConcept101~\cite{custom_diffusion} dataset.

\paragraph{More Results on Abstract Concepts.}

We present further examples demonstrating our method's capability to capture abstract concepts that are difficult to describe verbally. We utilized datasets from Freepik~\cite{Freepik} and ActionBench~\cite{actionbench} for these experiments. The results in Figure~\ref{fig:additional_abstract_concept} demonstrate that our learned token effectively captures abstract semantics, color palettes, and artistic styles from the given images.

\end{document}